\pdfoutput=1
\documentclass[letterpaper]{article}
\usepackage[totalwidth=480pt, totalheight=680pt]{geometry}

\usepackage{mathtools} 
\usepackage{booktabs} 
\usepackage{tikz} 

\title{Adaptivity Complexity for Causal Graph Discovery}
\author{
Davin Choo\thanks{Equal contribution}\\
National University of Singapore\\
\texttt{davin@u.nus.edu}
\and
Kirankumar Shiragur\footnotemark[1]\\
Broad Institute of MIT and Harvard\\
\texttt{shiragur@stanford.edu}
}
\date{}

\usepackage{algorithm}
\usepackage{algorithmicx}
\usepackage{algpseudocode}
\usepackage{subcaption}
\usepackage{xcolor}
\definecolor{darkgreen}{rgb}{0,0.5,0}
\usepackage{hyperref}
\hypersetup{
    unicode=false,          
    colorlinks=true,        
    linkcolor=red,          
    citecolor=darkgreen,    
    filecolor=magenta,      
    urlcolor=blue           
}
\usepackage[capitalize, nameinlink]{cleveref}

\usepackage{amsthm}
\usepackage{thm-restate}
\theoremstyle{plain}
\newtheorem{theorem}{Theorem}
\newtheorem{lemma}[theorem]{Lemma}
\theoremstyle{definition}
\newtheorem{definition}[theorem]{Definition}

\usetikzlibrary{calc, graphs, graphs.standard, shapes, arrows, arrows.meta, positioning, decorations.pathreplacing, decorations.markings, decorations.pathmorphing, fit, matrix, patterns, shapes.misc, tikzmark}

\newcommand{\cA}{\mathcal{A}}
\newcommand{\cB}{\mathcal{B}}
\newcommand{\cC}{\mathcal{C}}
\newcommand{\cE}{\mathcal{E}}
\newcommand{\cO}{\mathcal{O}}
\newcommand{\cI}{\mathcal{I}}
\newcommand{\skel}{\mathrm{skel}}

\begin{document}
\maketitle

\begin{abstract}
Causal discovery from interventional data is an important problem, where the task is to design an interventional strategy that learns the hidden ground truth causal graph $G(V,E)$ on $|V| = n$ nodes while minimizing the number of performed interventions.
Most prior interventional strategies broadly fall into two categories: non-adaptive and adaptive.
Non-adaptive strategies decide on a single fixed set of interventions to be performed while adaptive strategies can decide on which nodes to intervene on sequentially based on past interventions.
While adaptive algorithms may use exponentially fewer interventions than their non-adaptive counterparts, there are practical concerns that constrain the amount of adaptivity allowed.
Motivated by this trade-off, we study the problem of $r$-adaptivity, where the algorithm designer recovers the causal graph under a total of $r$ sequential rounds whilst trying to minimize the total number of interventions.
For this problem, we provide a $r$-adaptive algorithm that achieves $O(\min\{r,\log n\} \cdot n^{1/\min\{r,\log n\}})$ approximation with respect to the verification number, a well-known lower bound for adaptive algorithms. Furthermore, for every $r$, we show that our approximation is tight.
Our definition of $r$-adaptivity interpolates nicely between the non-adaptive ($r=1$) and fully adaptive ($r=n$) settings where our approximation simplifies to $O(n)$ and $O(\log n)$ respectively, matching the best-known approximation guarantees for both extremes.
Our results also extend naturally to the bounded size interventions.
\end{abstract}

\section{Introduction}
\label{sec:intro}

Learning causal relationships from data is a fundamental problem that has found applications across various scientific disciplines, including biology \cite{king2004functional,cho2016reconstructing,tian2016bayesian, sverchkov2017review,rotmensch2017learning,pingault2018using, de2019combining}, epidemiology, philosophy \cite{reichenbach1956direction,woodward2005making,eberhardt2007interventions}, and econometrics \cite{hoover1990logic,rubin2006estimating}.
Directed acyclic graphs (DAGs) are a popular choice to model causal relationships and it is well known that using observational data, the causal structure can only be learned up to its Markov equivalence class (MEC) and additional assumptions or interventional data is required for the recovery task.
Here, we focus our attention on causal discovery using interventions.
As interventions are often costly, our objective is to minimize interventions during the recovery process.

There is a rich literature on causal discovery from interventional data, and causal discovery algorithms can be broadly classified into two categories: adaptive \cite{shanmugam2015learning,greenewald2019sample,squires2020active,choo2022verification,choo2023subset} versus non-adaptive \cite{eberhardt2005number,eberhardt2006n,eberhardt2010causal,hu2014randomized}.
Given an essential graph, \emph{non-adaptive} algorithms have to decide beforehand a collection of interventions such that \emph{any} plausible causal graph can be recovered while \emph{adaptive} algorithms can decide on interventions sequentially while using information gleaned from past interventions.

Adaptive algorithms are powerful and the interventional cost of an optimal adaptive algorithm can even be exponentially better than any non-adaptive algorithms in some cases\footnote{On tree causal graphs, $\Omega(n)$ non-adaptive interventions are needed in the worst case while $\cO(\log n)$ adaptive ones suffice.}.
However, the sequential nature of adaptive algorithms fundamentally hinders parallelization \cite{dean2008approximating, balkanski2018adaptive} and may even be practically infeasible, e.g.\ hard constraints like timed deadlines may restrict how many rounds of adaptivity one can afford to do whilst minimizing the number of interventions.

The study of adaptivity is natural and has been studied across a wide spectrum of problems in computer science and statistics including parallel algorithms \cite{valiant1975parallelism, cole1988parallel,braverman2016parallel}, communication complexity \cite{papadimitriou1982communication,duris1984lower,nisan1991rounds,alon2015welfare}, multi-armed bandits \cite{agarwal2017learning}, sparse recovery problems \cite{malioutov2008compressed,haupt2009adaptive,indyk2011power}, knapsack \cite{dean2008approximating}, submodular optimization \cite{balkanski2018adaptive}, stochastic set covering \cite{goemans2006stochastic}, stochastic probing \cite{gupta2016algorithms,gupta2017adaptivity}, local search algorithms \cite{branzei2022query}, and many more \cite{scarlett2018noisy,canonne2018adaptivity,raskhodnikova2006note}.
Typically, one can obtain a lower objective cost if more rounds of adaptivity are allowed and researchers have been trying to characterize the trade-off between cost and adaptivity in various problems.

Beyond being a fundamental question in theory, understanding adaptivity is also practically motivated in the problem of causal discovery.
Consider a scientific institute where lab technicians carry out experiments suggested by their scientist colleagues to make progress towards a common research goal.
A round of interaction involves the experimentalists discussing with the scientists on what experiment(s) should be performed next, where each round of interaction may incur undesirable coordination overheads.
Since the institute has enough resources to run multiple (but bounded) number of experiments simultaneously, a batch of experiments is typically proposed in a single round of interaction.
In this work, we formally model the interactions between the two parties through the framework of adaptivity in hopes that our theoretical contributions will yield practical insights, e.g.\ give guidance on how the two parties should interact.
Let us now formally define our problem setup.

\paragraph{Problem setup}
Given $r \geq 1$ adaptivity rounds and a partially oriented observational essential graph of an underlying causal graph $G^* = (V,E)$, adaptively design\footnote{The decision of $\cI_i$ may depend on outcomes after recovering arc orientations from intervening on $\cI_1, \ldots, \cI_{i-1}$.} intervention sets $\cI_1, \ldots, \cI_r \subseteq 2^V$ such that all arc orientations in $G^*$ will be recovered after intervening on them while minimizing the total number of interventions performed.

Formally, we require $\cE_{\cI_1 \cup \ldots \cup \cI_r}(G^*) = G^*$ while minimizing $\cup_{i=1}^r |\cI_i|$, where $\cE_{\cI}(G^*)$ denotes the partially oriented interventional essential graph after performing interventions $\cI$.
Furthermore, each intervention $I \in \cI_i$ in any intervention set $\cI_i$ has size $|I| \leq k$.
Atomic interventions when $k = 1$ and bounded size interventions when $k > 1$.

\subsection{Contributions}

Under standard assumptions of causal sufficiency, faithfulness and infinite sample regime\footnote{These assumptions are common across all/most theoretical causal graph discovery works and one may also interpret these assumptions as having access to a conditional independence oracle.}, we provide a $r$-adaptive search algorithm that recovers the ground truth causal graph by performing at most $\cO(\min\{r,\log n\} \cdot n^{1/\min\{r,\log n\}} \cdot \nu_1(G^*))$ atomic interventions, where $\nu_1(G^*)$ is the atomic verification number of $G^*$ \cite{squires2020active,choo2022verification,choo2023subset}\footnote{Given a MEC of an unknown ground truth causal graph $G^*$ and a graph $G$ from the MEC, the goal of the verification problem is determining whether $G$ is $G^*$. By plugging in $G$ with $G^*$ in the verification problem, we see that the optimal solution to the verification is a natural lower bound for the search problem.}, a natural lower bound for the adaptive algorithms.
See \cref{def:verifying-set-and-verification-number} for definition of $\nu_1(G^*)$ and $\nu_k(G^*)$, the bounded size analog of $\nu_1(G^*)$.
To the best of our knowledge, this is the first work to formalize the trade-offs between sequentiality and parallelism by studying adaptivity in the context of causal graph discovery.

\begin{restatable}[Atomic upper bound]{theorem}{atomicupperbound}
\label{thm:atomic-upper-bound}
Let $\cE(G^*)$ be the observational essential graph of an underlying causal DAG $G^*$ on $n$ nodes and $m$ edges.
There is a $r$-adaptive algorithm (\cref{alg:adaptive-search}) that uses $\cO(\min\{r,\log n\} \cdot n^{1/\min\{r,\log n\}} \cdot \nu_1(G^*))$ atomic interventions to recover $G^*$ from $\cE(G^*)$.
Ignoring the time spent implementing the actual interventions, \cref{alg:adaptive-search} runs in $\cO(n^{1+1/r} \cdot (n+m))$ time.
\end{restatable}

When $r = 1$, the upper bound becomes $\cO(n)$, which is worst case optimal since $\Omega(n)$ non-adaptive interventions are necessary when the given essential graph is a path on $n$ vertices.
Meanwhile, when $r = \log n$, the upper bound becomes $\cO(\log n \cdot \nu_1(G^*))$, matching the upper bound guarantees of \cite{choo2022verification}.
Again, this is worst case optimal when the given essential graph is a path on $n$ vertices.

In fact, for any $r$, our approximation factor is tight.
We formally show this in the following lower bound result.

\begin{restatable}[Atomic worst case]{theorem}{atomicworstcase}
\label{thm:atomic-worst-case}
In the worst case, any $r$-adaptive algorithm needs to use at least
$\Omega(\min\{r,\log n\} \cdot n^{1/\min\{r,\log n\}} \cdot \nu_1(G^*))$ atomic interventions against an adaptive adversary.
\end{restatable}

We also extend our upper bound results to accommodate bounded size interventions, where each intervention can involve up to $k$ vertices, for some pre-determined bound $k \geq 1$; atomic interventions are a special case of $k = 1$.

\begin{restatable}[Bounded upper bound]{theorem}{boundedupperbound}
\label{thm:bounded-upper-bound}
Let $\cE(G^*)$ be the observational essential graph of an underlying causal DAG $G^*$ on $n$ nodes.
There is a polynomial time $r$-adaptive algorithm that uses $\cO(\min\{r,\log n\} \cdot n^{1/\min\{r,\log n\}} \cdot \log k \cdot \nu_k(G^*))$ bounded sized interventions to recover $G^*$ from $\cE(G^*)$, where each intervention involves at most $k > 1$ vertices.
\end{restatable}

While the approximation ratio is worse than \cref{thm:atomic-upper-bound}, note that we are comparing against $\nu_k(G^*)$.
Since using bounded size interventions typically translates to using a smaller number of interventions performed as $k$ increases, $\nu_k(G^*)$ could roughly be $k$ times smaller than $\nu_1(G^*)$.

\paragraph{Remark}
Since non-adaptive results and the upper bound guarantees of \cite{choo2022verification} already matches our upper bounds for regime of $r \geq \log n$, it suffices for us to design and analyze algorithms for regime of $1 < r < \log n$.

\subsection{Outline}

After giving preliminaries and related work in \cref{sec:prelims}, we slowly build up intuition towards our main results.
We first solve the problem using atomic interventions when the input graph is a path or a tree (\cref{sec:warmup}) before solving the problem in full generality on any graph input, and also generalizing to the setting of bounded size interventions (\cref{sec:full-generality}).
We show how our algorithm performs in practice on synthetic graphs in \cref{sec:experiments} and conclude with some interesting future work directions in \cref{sec:conclusion}.
For a cleaner exposition, some details are deferred to the appendix.

\section{Preliminaries}
\label{sec:prelims}

We write $\{1, \ldots, n\}$ as $[n]$ and use standard asymptotic notations such as $\cO(\cdot)$ and $\Omega(\cdot)$.
$A \;\dot\cup\; B$ refers to the union of two disjoint sets $A$ and $B$.
Logarithms are in base 2.

\subsection{Graph notations}

Let $G = (V,E)$ be a graph on $|V| = n$ vertices.
We use $V(G)$, $E(G)$ and $A(G) \subseteq E(G)$ to denote its vertices, edges, and oriented arcs respectively.
The graph $G$ is said to be directed or fully oriented if $A(G) = E(G)$, and partially oriented otherwise.
For any two vertices $u,v \in V$, we write $u \sim v$ if these vertices are connected in the graph and $u \not\sim v$ otherwise.
To specify the arc directions, we use $u \to v$ or $u \gets v$.
We use $G[V']$ to denote the vertex-induced subgraph for any subset $V' \subseteq V$ of vertices.

A \emph{clique} is a graph where $u \sim v$ for any pair of vertices $u,v \in V$.
A \emph{maximal clique} is an vertex-induced subgraph of a graph that is a clique and ceases to be one if we add any other vertex to the subgraph.
For an undirected graph $G$, $\omega(G)$ refers to the size of its maximum clique.

The \emph{skeleton} $skel(G)$ of a (partially oriented) graph $G$ is the underlying graph where all edges are made undirected.
A \emph{v-structure} refers to three distinct vertices $u,v,w \in V$ such that $u \to v \gets w$ and $u \not\sim w$.
A simple cycle is a sequence of $k \geq 3$ vertices where $v_1 \sim v_2 \sim \ldots \sim v_k \sim v_1$.
The cycle is partially directed if at least one of the edges is directed and all directed arcs are in the same direction along the cycle.
A partially directed graph is a \emph{chain graph} if it contains no partially directed cycle.
In the undirected graph $G[E \setminus A]$ obtained by removing all arcs from a chain graph $G$, each connected component in $G[E \setminus A]$ is called a \emph{chain component}.
We use $CC(G)$ to denote the set of chain components, where each $H \in CC(G)$ is a subgraph of $G$ and $V = \dot\cup_{H \in CC(G)} V(H)$.

\subsection{Chordal graphs}

An undirected graph is chordal if every cycle of length at least 4 has an edge connecting two non-adjacent vertices of the cycle.
There are many known characterizations and properties of chordal graphs; see \cite{blair1993introduction} for an introduction.
One of the most common characterization is the following: A graph $G$ is chordal if and only if perfect elimination ordering (PEO)\footnote{An ordering $\sigma$ is a PEO if for $1 \leq i \leq n$, the node-induced subgraph $G[\{v_1, \ldots, v_{i-1}\} \cap N(v_i)]$ on $v_i$'s neighbors is a clique.} exists \cite{fulkerson1965incidence}.
Furthermore, a PEO can be computed in linear time via lexicographic BFS \cite{rose1976algorithmic} and can used to prove the following lemma which implies that a chordal graph on $n$ nodes has at most $n$ maximal cliques.

\begin{lemma}[\cite{fulkerson1965incidence}; Lemma 6 in \cite{blair1993introduction}]
The set of maximal cliques of a graph $G$ is precisely the sets $\{v_i\} \cup (\{v_{i+1}, \ldots, v_n\} \cap N(v_i))$ for which $\{v_i\} \cup (\{v_{i+1}, \ldots, v_n\} \cap N(v_i))$ is \emph{not} in $\{v_j\} \cup (\{v_{j+1}, \ldots, v_n\}\cap N(v_j))$ for any vertex $v_j$ with $j < i$.
\end{lemma}

It is known that chordal graphs have a clique tree representation.
One way to construct a clique tree $T_G$ from a chordal graph $G$ in polynomial time is via the ``maximum-weight spanning tree property'':
Let the vertex set of $T_G$ be all maximal cliques of $G$, assign edge weight as the size of intersection between every pair of maximal cliques, and then compute the \emph{maximum} weight spanning tree.

The following result is one of the many useful properties of clique trees which we exploit in our algorithm later.

\begin{lemma}[Lemma 5 of \cite{blair1993introduction}]
\label{lem:prelim-lemma-5}
Let $T_G = (K, S)$ be the clique tree of a chordal graph $G = (V, E)$.
For any two adjacent maximal cliques $K_i$ and $K_j$ in $T_G$, let $T_i$ and $T_j$ be the subtrees obtained by removing the edge $\{K_i, K_j\}$ from $T_G$.
Then, vertices $v_i$ and $v_j$ are disconnected in $G[V \setminus (V(K_i) \cap V(K_j))]$.
\end{lemma}

\subsection{Causal graph basics}

Directed acyclic graphs (DAGs), a special case of chain graphs where \emph{all} edges are directed, are commonly used as graphical causal models \cite{pearl2009causality} where vertices represents random variables and the joint probability density $f$ factorizes according to the Markov property:
$
f(v_1, \ldots, v_n) = \prod_{i=1}^n f(v_i \mid pa(v_i))
$, where $pa(v_i)$ is the values taken by $v_i$'s parents in the DAG.

For any DAG $G$, we denote its \emph{Markov equivalence class} (MEC) by $[G]$ and \emph{essential graph} by $\cE(G)$.
Two graphs are Markov equivalent if and only if they have the same skeleton and v-structures \cite{verma1990,andersson1997characterization}.
DAGs in the same MEC have the same skeleton;
Essential graph is a partially directed graph such that an arc $u \to v$ is directed if $u \to v$ in \emph{every} DAG in MEC $[G]$, and an edge $u \sim v$ is undirected if there exists two DAGs $G_1, G_2 \in [G]$ such that $u \to v$ in $G_1$ and $v \to u$ in $G_2$.

An \emph{intervention} $S \subseteq V$ is an experiment where all variables $s \in S$ are forcefully set to some value, independent of the underlying causal structure.
An intervention is \emph{atomic} if $|S| = 1$ and \emph{bounded} if $|S| \leq k$ for some $k>0$; observational data is a special case where $S = \emptyset$.
The effect of interventions is formally captured by Pearl's do-calculus \cite{pearl2009causality}.
We call any $\cI \subseteq 2^V$ an \emph{intervention set}: an intervention set is a set of interventions where each intervention corresponds to a subset of variables.
An \emph{ideal intervention} on $S \subseteq V$ in $G$ induces an interventional graph $G_S$ where all incoming arcs to vertices $v \in S$ are removed \cite{eberhardt2012number}.
It is known that intervening on $S$ allows us to infer the edge orientation of any edge cut by $S$ and $V \setminus S$ \cite{eberhardt2007causation,hyttinen2013experiment,hu2014randomized,shanmugam2015learning,kocaoglu2017cost}.

For ideal interventions, an $\cI$-essential graph $\cE_{\cI}(G)$ of $G$ is the essential graph representing the Markov equivalence class of graphs whose interventional graphs for each intervention is Markov equivalent to $G_S$ for any intervention $S \in \cI$.
There are several known properties about $\cI$-essential graph properties (e.g.\ see \cite{hauser2012characterization,hauser2014two}).
For instance, every $\cI$-essential graph is a chain graph with chordal chain components; this includes the case of $S = \emptyset$.
Also, orientations in one chain component do not affect orientations in other components.
Thus, to fully orient any essential graph $\cE(G^*)$, it is necessary and sufficient to orient every chain component in $\cE(G^*)$.

A \emph{verifying set} $\cI$ for a DAG $G \in [G^*]$ is an intervention set that fully orients $G$ from $\cE(G^*)$, possibly with repeated applications of Meek rules (see Appendix A).
In other words, for any graph $G = (V,E)$ and any verifying set $\cI$ of $G$, we have $\cE_{\cI}(G)[V'] = G[V']$ for \emph{any} subset of vertices $V' \subseteq V$.
Furthermore, if $\cI$ is a verifying set for $G$, then $\cI \cup S$ is also a verifying set for $G$ for any additional intervention $S \subseteq V$.
While DAGs may have multiple verifying sets in general, we are often interested in finding one with minimum size.

\begin{definition}[Verifying set and verifying number]
\label{def:verifying-set-and-verification-number}
An intervention set $\cI$ is called a verifying set for a DAG $G^*$ if $\cE_{\cI}(G^*) = G^*$.
$\cI$ is a \emph{minimum size verifying set} if $\cE_{\cI'}(G^*) \neq G^*$ for any $|\cI'| < |\cI|$.
The \emph{verification number} $\nu_k(G^*)$ denotes the size of the minimum size verifying set of $G^*$ when each intervention has size at most $k \geq 1$.
\end{definition}

Recently, \cite{choo2022verification} proved two fundamental results with respect to verification number.
\cref{thm:verification-characterization} characterizes verification number of any given causal DAG while \cref{thm:neuripssearch} gives an adaptive search algorithm that is competitive to the verification number of the underlying causal DAG.

\begin{theorem}[\cite{choo2022verification}]
\label{thm:verification-characterization}
Fix an essential graph $\cE(G^*)$ and $G \in [G^*]$.
An atomic intervention set $\cI$ is a minimal sized verifying set for $G$ if and only if $\cI$ is a minimum vertex cover of covered edges $\cC(G)$ of $G$.
A minimal sized atomic verifying set can be computed in polynomial time since the edge-induced subgraph on $\cC(G)$ is a forest.
\end{theorem}

\begin{theorem}[\cite{choo2022verification}]
\label{thm:neuripssearch}
Fix an unknown underlying DAG $G^*$.
Given an essential graph $\cE(G^*)$ and intervention set bound $k \geq 1$, there is a deterministic polynomial time algorithm that computes an intervention set $\cI$ adaptively such that $\cE_{\cI}(G^*) = G^*$, and $|\cI|$ has size\\
1. $\cO(\log(n) \cdot \nu_1(G^*))$ when $k = 1$\\
2. $\cO(\log(n) \cdot \log (k) \cdot \nu_k(G^*))$ when $k > 1$.
\end{theorem}

To obtain a competitive bound with respect to $\nu_1(G^*)$, \cite{choo2022verification} proved a stronger (but non-computable) lower bound (\cref{thm:neuripssearch}) on $\nu_1(G^*)$.
We will rely on this lemma to show a competitive bound for our algorithm later.

\begin{restatable}{lemma}{strengthenedlb}
\label{lem:strengthened-lb}
For any causal DAG $G^*$,
\[
\nu_1(G^*) \geq \max_{\cI \subseteq V} \sum_{H \in CC(\cE_{\cI}(G^*))} \left\lfloor \frac{\omega(H)}{2} \right\rfloor
\]
\end{restatable}

For any intervention set $\cI \subseteq 2^V$, we write $R(G, \cI) \subseteq E$ to mean the set of oriented arcs in the $\cI$-essential graph of a DAG $G$, and define $G^{\cI} = G[E \setminus R(G,\cI)]$ as the \emph{fully directed} subgraph DAG induced by the \emph{unoriented arcs} in $G$, where $G^{\emptyset}$ is the graph obtained after removing all the oriented arcs in the observational essential graph due to v-structures.
The next result explains why it suffices to study causal graph discovery via interventions on causal graphs without v-structures: since $R(G,\cI) = R(G^{\emptyset},\cI) \;\dot\cup\; R(G, \emptyset)$, any oriented arcs in the observational graph can be removed \emph{before performing any interventions} as the optimality of the solution is unaffected.

\begin{theorem}[\cite{choo2023subset}]
\label{thm:can-remove-oriented}
For any DAG $G = (V,E)$ and intervention sets $\cA, \cB \subseteq 2^V$,
\[
R(G,\cA \cup \cB)
= R(G^{\cA},\cB) \;\dot\cup\ R(G^{\cB},\cA) \;\dot\cup\; (R(G,\cA) \cap R(G,\cB))
\]
\end{theorem}

We will also borrow the following definition of \emph{relevant nodes} from \cite{choo2023subset}.

\begin{definition}[Relevant nodes]
Fix a DAG $G^* = (V,E)$ and arbitrary subset $V' \subseteq V$.
For any intervention set $\cI \subseteq V$ and resulting interventional essential graph $\cE_{\cI}(G^*)$, we define the \emph{relevant nodes} $\rho(\cI, V') \subseteq V'$ as the set of nodes within $V'$ that is adjacent to some unoriented arc within the node-induced subgraph $\cE_{\cI}(G^*)[V']$.
\end{definition}

\subsection{Related work}

There are two broad classes of causal graph discovery algorithms.
Given the observational essential graph, non-adaptive algorithms need to recover any underlying causal graph using a single fixed set of interventions while adaptive algorithms are allowed to adapt their interventional decisions based on the outcomes of earlier interventions.

\paragraph{Non-adaptive search}
Separating systems are the central mathematical objects for non-adaptive intervention design.
Roughly speaking, a separating system on a set of elements is a collection of subsets such that for every pair of elements from the set, there exists at least one subset which contains exactly one element from the pair.
Instead of all pairs of elements, let us consider the (typically smaller) $G$-separating system for a given undirected graph $G$.

\begin{definition}[$G$-separating system; Definition 3 of \cite{kocaoglu2017cost}]
Given an undirected graph $G = (V,E)$, a set of subsets $\cI \subseteq 2^V$ is a $G$-separating system if for every edge $\{u,v\} \in E$, there exists $I \in \cI$ such that either ($u \in I_i$ and $v \not\in I_i$) or ($u \not\in I_i$ and $v \in I_i$).
\end{definition}

To simplify notation, we write $G$-separating system to refer to $\skel(G)$-separation system or $\skel(\cE(G))$-separation system, for any causal DAG $G$.
It is known that the optimal non-adaptive intervention set to learn a moral DAG $G^*$ is a $G^*$-separating system \cite{kocaoglu2017cost}.

\begin{theorem}[Theorem 1 of \cite{kocaoglu2017cost}]
\label{thm:g-separating-system}
For any undirected graph $G$, an intervention set $\cI$ learns \emph{every} possible causal graph $D$ with $\skel(D) = G$ if and only if $\cI$ is a $G$-separating system.
\end{theorem}

\paragraph{Adaptive search}
There exists essential graphs\footnote{If the essential graph is an undirected path on $n$ vertices, then a $G$-sparating system has size $\Omega(n)$ while adaptive search only requires $\cO(\log n)$ atomic interventions by ``binary search''.} where non-adaptive interventions require exponentially more interventions than if one could use adaptive interventions.

Adaptive search has been studied for special graph classes by \cite{shanmugam2015learning,greenewald2019sample,squires2020active}.
More recently, \cite{choo2022verification} showed that $\cO(\log n \cdot \nu_1(G^*))$ atomic inteventions suffices to fully recover for any general causal graph $G^*$ on $n$ nodes while \cite{choo2023subset} showed that $\cO(\log |\rho(\emptyset, V(H))| \cdot \nu_1(G^*))$ atomic inteventions suffices to fully recover edge directions within a subgraph $H$ of $G^*$.
Both papers also gave results in terms of bounded size interventions where their guarantees incur an additional $\cO(\log k)$ factor when comparing to $\nu_k(G^*)$ by invoking \cref{lem:labelling-lemma} suitably.

\begin{lemma}[Lemma 1 of \cite{shanmugam2015learning}]
\label{lem:labelling-lemma}
Let $(n,k,a)$ be parameters where $k \leq n/2$.
There is a polynomial time labeling scheme that produces distinct $\ell$ length labels for all elements in $[n]$ using letters from the integer alphabet $\{0\} \cup [a]$ where $\ell = \lceil \log_a n \rceil$.
In every label index, any integer letter is used at most $\lceil n/a \rceil$ times.
This labelling scheme is a separating system: for any $i,j \in [n]$, there exists some digit $d \in [\ell]$ where the labels of $i$ and $j$ differ.
\end{lemma}

\section{Warmup: Paths and Trees}
\label{sec:warmup}

When the essential graph $\cE(G^*)$ is a path or tree, there are $n$ possible DAGs, each associated with setting a node as a root and orienting all edges away from it.
It is known that $\nu(G^*) = 1$ because the covered edges for each DAG are precisely the edges incident to the hidden root (e.g.\ see \cite{greenewald2019sample,choo2022verification}).

In this section, we investigate how to optimally solve with $r$-adaptivity on these special classes of graphs.

\subsection{Paths}
\label{sec:paths}

When the adaptivity parameter $r = 1$, any $1$-adaptive algorithm behaves \emph{exactly} like a non-adaptive algorithm.
Thus, it is necessary and sufficient to intervene on a $G^*$-seperating system, and such a system on a path has size $\Theta(n)$.

Meanwhile, when the adaptive parameter $r = 2$, we already see an interesting trade-off occurring: \emph{how much (and how) should we intervene now versus later?}
Our next lemma tells us how to balance the number of interventions done in the first and second adaptive rounds in a worst case fashion.

\begin{lemma}
\label{lem:path-r-2}
Suppose $\cE(G^*)$ is a path on $n$ vertices.
There is a $2$-adaptive algorithm that uses at most $\cO(\sqrt{n})$ interventions in total.
In the worst case, $\Omega(\sqrt{n})$ is necessary for any $2$-adaptive algorithm.
\end{lemma}
\begin{proof}
Without loss of generality, up to the inclusion of floors and ceilings, let us assume that $n$ is a square number.

Suppose the vertices on the path are labelled $v_1, v_2, \ldots, v_n$ where $v_i \sim v_{i+1}$ for $1 \leq i \leq n-1$.

In the first round, we intervene on $\sqrt{n}$ evenly spaced vertices: $v_{\sqrt{n}}, v_{2 \sqrt{n}}, v_{3 \sqrt{n}}, \ldots, v_{n}$.
After applying Meek's rule R1, at most one segment of $v_{i \cdot \sqrt{n}}, v_{i \cdot \sqrt{n} + 1}, v_{i \cdot \sqrt{n} + 2}, \ldots, v_{(i+1) \cdot \sqrt{n}}$ will remain unoriented.
That is, the total number of relevant vertices is at most $\sqrt{n}$.
In the second round, we intervene on all these relevant vertices, incurring a total of $\cO(\sqrt{n})$ interventions.

For the worst case lower bound, consider any arbitrary $2$-adaptive algorithm $A$.
If $A$ makes strictly less than $\sqrt{n}$ interventions in the first round, then there exists a consecutive sequence of $\sqrt{n}$ vertices $v_i, v_{i+1}, \ldots, v_{i+\sqrt{n}}$ which is not intervened by $A$.
If the root node of the path was $v_i$, then this entire segment $v_i \sim v_{i+1} \sim \ldots \sim v_{i+\sqrt{n}}$ remains unoriented after the first round of interventions.
Thus, in the second round, $A$ needs to at least intervene on a separating system of this segment, which has a size at least $\Omega(\sqrt{n})$.
Therefore, in the worst case, any $2$-adaptive algorithm needs to perform at least $\Omega(\sqrt{n})$ number of interventions to orient $\cE(G^*)$ when it is a path on $n$ vertices.
\end{proof}

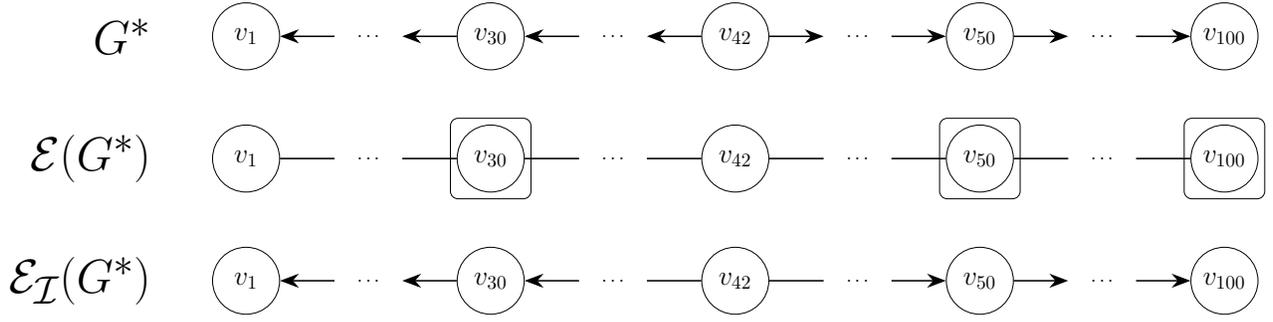
\begin{figure}[htb]
\centering
\resizebox{\linewidth}{!}{%
\begin{tikzpicture}
%
%
\node[draw, circle, minimum size=35pt, inner sep=2pt] at (0,0) (v1) {\Large $v_1$};
\node[minimum size=35pt, inner sep=2pt, right=of v1] (v2) {$\ldots$};
\node[draw, circle, minimum size=35pt, inner sep=2pt, right=of v2] (v3) {\Large $v_{30}$};
\node[minimum size=35pt, inner sep=2pt, right=of v3] (v4) {$\ldots$};
\node[draw, circle, minimum size=35pt, inner sep=2pt, right=of v4] (v5) {\Large $v_{42}$};
\node[minimum size=35pt, inner sep=2pt, right=of v5] (v6) {$\ldots$};
\node[draw, circle, minimum size=35pt, inner sep=2pt, right=of v6] (v7) {\Large $v_{50}$};
\node[minimum size=35pt, inner sep=2pt, right=of v7] (v8) {$\ldots$};
\node[draw, circle, minimum size=35pt, inner sep=2pt, right=of v8] (v9) {\Large $v_{100}$};
\draw[thick, {Stealth[scale=1.5]}-] (v1) -- (v2);
\draw[thick, {Stealth[scale=1.5]}-] (v2) -- (v3);
\draw[thick, {Stealth[scale=1.5]}-] (v3) -- (v4);
\draw[thick, {Stealth[scale=1.5]}-] (v4) -- (v5);
\draw[thick, -{Stealth[scale=1.5]}] (v5) -- (v6);
\draw[thick, -{Stealth[scale=1.5]}] (v6) -- (v7);
\draw[thick, -{Stealth[scale=1.5]}] (v7) -- (v8);
\draw[thick, -{Stealth[scale=1.5]}] (v8) -- (v9);
\node[left=of v1] {\Huge $G^*$};

%
%
\node[draw, circle, minimum size=35pt, inner sep=2pt, below=of v1] (ev1) {\Large $v_1$};
\node[minimum size=35pt, inner sep=2pt, right=of ev1] (ev2) {$\ldots$};
\node[draw, circle, minimum size=35pt, inner sep=2pt, right=of ev2] (ev3) {\Large $v_{30}$};
\node[minimum size=35pt, inner sep=2pt, right=of ev3] (ev4) {$\ldots$};
\node[draw, circle, minimum size=35pt, inner sep=2pt, right=of ev4] (ev5) {\Large $v_{42}$};
\node[minimum size=35pt, inner sep=2pt, right=of ev5] (ev6) {$\ldots$};
\node[draw, circle, minimum size=35pt, inner sep=2pt, right=of ev6] (ev7) {\Large $v_{50}$};
\node[minimum size=35pt, inner sep=2pt, right=of ev7] (ev8) {$\ldots$};
\node[draw, circle, minimum size=35pt, inner sep=2pt, right=of ev8] (ev9) {\Large $v_{100}$};
\draw[thick] (ev1) -- (ev2);
\draw[thick] (ev2) -- (ev3);
\draw[thick] (ev3) -- (ev4);
\draw[thick] (ev4) -- (ev5);
\draw[thick] (ev5) -- (ev6);
\draw[thick] (ev6) -- (ev7);
\draw[thick] (ev7) -- (ev8);
\draw[thick] (ev8) -- (ev9);
\node[left=of ev1] {\Huge $\cE(G^*)$};
\node[draw, rounded corners, fit=(ev3)]{};
\node[draw, rounded corners, fit=(ev7)]{};
\node[draw, rounded corners, fit=(ev9)]{};

%
%
\node[draw, circle, minimum size=35pt, inner sep=2pt, below=of ev1] (iv1) {\Large $v_1$};
\node[minimum size=35pt, inner sep=2pt, right=of iv1] (iv2) {$\ldots$};
\node[draw, circle, minimum size=35pt, inner sep=2pt, right=of iv2] (iv3) {\Large $v_{30}$};
\node[minimum size=35pt, inner sep=2pt, right=of iv3] (iv4) {$\ldots$};
\node[draw, circle, minimum size=35pt, inner sep=2pt, right=of iv4] (iv5) {\Large $v_{42}$};
\node[minimum size=35pt, inner sep=2pt, right=of iv5] (iv6) {$\ldots$};
\node[draw, circle, minimum size=35pt, inner sep=2pt, right=of iv6] (iv7) {\Large $v_{50}$};
\node[minimum size=35pt, inner sep=2pt, right=of iv7] (iv8) {$\ldots$};
\node[draw, circle, minimum size=35pt, inner sep=2pt, right=of iv8] (iv9) {\Large $v_{100}$};
\draw[thick, {Stealth[scale=1.5]}-] (iv1) -- (iv2);
\draw[thick, {Stealth[scale=1.5]}-] (iv2) -- (iv3);
\draw[thick, {Stealth[scale=1.5]}-] (iv3) -- (iv4);
\draw[thick] (iv4) -- (iv5);
\draw[thick] (iv5) -- (iv6);
\draw[thick, -{Stealth[scale=1.5]}] (iv6) -- (iv7);
\draw[thick, -{Stealth[scale=1.5]}] (iv7) -- (iv8);
\draw[thick, -{Stealth[scale=1.5]}] (iv8) -- (iv9);
\node[left=of iv1] {\Huge $\cE_{\cI}(G^*)$};
\end{tikzpicture}
}
\caption{Path example with $n=100$ with $v_{42}$ as the hidden source. After intervening on $\cI = \{v_{10}, v_{20}, \ldots, v_{100}\}$ atomically and applying Meek rules, only the segment $v_{41} \sim v_{42} \sim \ldots \sim v_{49}$ remains unoriented.}
\label{fig:path-example}
\end{figure}

The key technical algorithmic idea in \cref{lem:path-r-2} is the strategy of ``balanced partitioning'', which balances the worst case size of the largest possible unoriented component after a round of intervention.
See \cref{fig:path-example} for an example.

Interestingly, we can generalize this ``balanced partitioning'' strategy to larger values of adaptivity parameter $r$ as follows:
Suppose we perform $L$ interventions per round in the first $r-1$ rounds and then intervene on the $G$-separating system on the remaining relevant vertices in the final round, where $L$ is an integer that depends on $r$ which we define later.
In each of the $r-1$ rounds, we will choose the $L$ vertices judiciously by a ``balanced partitioning''.
If we further insist that the final $G$-separating system has size at most $L$, then we get a recurrence relation $\frac{n}{(L+1)^{r-1}} \leq L$ while incurring a total of $\approx r \cdot L$ interventions.
We formally prove this next.

\begin{lemma}
\label{lem:path-k}
Suppose $\cE(G^*)$ is a path on $n$ vertices.
There is a $r$-adaptive algorithm that uses at most $\cO(r \cdot n^{1/r})$ interventions in total.
\end{lemma}
\begin{proof}
Observe that no matter how we intervene on a path, the remaining relevant vertices will form a subpath.

Let $L = \lceil n^{1/r} \rceil$.
In the first $r-1$ rounds, we will partition the remaining unoriented subpath into equal length segments.
For instance, if the subpath length is currently $\ell$, then we partition it into segments of length $\lceil \ell/(L+1) \rceil$ or $\lfloor \ell/(L+1) \rfloor$.
Thus, the length of the subpath with relevant vertices in the next round will be at most $\lceil \ell/(L+1) \rceil$.

After $r-1$ rounds, the length of the remaining unoriented subpath is at most (via repeated applications of \cref{lem:nested-divisions}):
\[
\left\lceil \left\lceil \left\lceil  \frac{n}{L+1} \right\rceil / (L+1) \right\rceil \ldots \right\rceil
\leq \ldots
\leq \left\lceil \frac{n}{(L+1)^{r-1}} \right\rceil
\]
Since $L = \lceil n^{1/r} \rceil$, we see that $\left\lceil \frac{n}{(L+1)^{r-1}} \right\rceil \leq L$.
Therefore, by intervening on all the remaining relevant vertices in the final $r$-th round, the total number of interventions performed will be $\cO(r \cdot L) \subseteq \cO(r \cdot n^{1/r})$.
\end{proof}

\begin{restatable}[See appendix for proof]{lemma}{nesteddivisions}
\label{lem:nested-divisions}
For positive integer $n$, and arbitrary real numbers $m,x$, we have
$
\left\lceil \frac{\left\lceil \frac{x}{m} \right\rceil}{n} \right\rceil
= \left\lceil \frac{x}{mn} \right\rceil
$.
\end{restatable}

\subsection{Trees}
\label{sec:trees}

While the high-level strategy employed in \cref{lem:path-k} also works for trees, how we perform the ``balanced partitioning'' is slightly different since the vertices are no longer arranged in a line.
Instead, we rely on \cref{alg:balanced-partitioning-trees} to partition a tree on $n$ nodes into subtrees using $L$ node removals such that each subtree has size at most $\lceil n/(L+1) \rceil$.

The correctness of \cref{alg:balanced-partitioning-trees} is proven in \cref{lem:tree-subroutine-works} and the resulting $r$-adaptive search result is given in \cref{lem:tree-r}.

\begin{algorithm}[htb]
\caption{Balanced partitioning on trees.}
\label{alg:balanced-partitioning-trees}
\begin{algorithmic}[1]
    \Statex \textbf{Input}: A tree $G = (V,E)$ with $|V| = n$, an integer $L$.
    \Statex \textbf{Output}: A subset of vertices $A \subseteq V$ such that $|A| \leq L$ and subtrees in $G[V \setminus A]$ have size at most $\lceil n/(L+1) \rceil$.
    \State Initialize $A \gets \emptyset$
    \While{tree $G = (V,E)$ has size $|V| > \lceil \frac{n}{L+1} \rceil$}
        \State Root $G$ arbitrarily.
        \State Compute size of subtrees $T_u$ at each node $u \in V$
        \If{there is a subtree $T_u$ of size $1 + \lceil \frac{n}{L+1} \rceil$}
            \State Add $u$ to $A$; Update $G \gets G[V \setminus V(T_u)]$.
        \Else
            \State Find a subtree $T_u$ of size $|V(T_u)| > 1 + \lceil \frac{n}{L+1} \rceil$, such that $|V(T_w)| \leq \lceil \frac{n}{L+1} \rceil$ for \emph{all} children $w$ of $u$.
            \State Add $u$ to $A$; Update $G \gets G[V \setminus V(T_u)]$.
        \EndIf
    \EndWhile
    \State \Return $A$
\end{algorithmic}
\end{algorithm}

\begin{lemma}
\label{lem:tree-subroutine-works}
Given a tree $G = (V,E)$ with $|V| = n$ and an integer $L \leq n$, \cref{alg:balanced-partitioning-trees} runs in polynomial time and returns a subset of vertices $A \subseteq V$ such that $|A| \leq L$ and subtrees in $G[V \setminus A]$ have size at most $\lceil n/(L+1) \rceil$.
\end{lemma}
\begin{proof}
We first prove the correctness then the running time.

\paragraph{Correctness}
Consider an arbitrary iteration of the while loop.
We will argue two things:
\begin{enumerate}
    \item The size of $G$ decreases by at least $\lceil \frac{n}{L+1} \rceil$.
    \item The newly pruned subtree(s) have size at most $\lceil \frac{n}{L+1} \rceil$.
\end{enumerate}
Since $n - L \cdot \lceil \frac{n}{L+1} \rceil \leq \lceil \frac{n}{L+1} \rceil$, the while loop terminates after at most $L$ iterations.
Thus, $|A| \leq L$ as we only add one vertex per iteration to $A$.

\emph{if-case:}
Since $|V(T_u)| = 1 + \lceil \frac{n}{L+1} \rceil$, removing $V(T_u)$ from $V$ decreases the size of $G$ by at least $\lceil \frac{n}{L+1} \rceil$.
Furthermore, the subtree pruned by $u$'s removal has size exactly $\lceil \frac{n}{L+1} \rceil$.

\emph{else-case:}
This case happens when \emph{no} subtrees has size exactly $1 + \lceil \frac{n}{L+1} \rceil$.
So, if a subtree has size strictly larger than $\lceil \frac{n}{L+1} \rceil$, then it must have size at least $2 + \lceil \frac{n}{L+1} \rceil$.

We first show how to find such a subtree $T_u$ in Line 8:
\begin{enumerate}
    \item Initialize $u$ as the root node.
    \item If all children $w$ of $u$ have subtree sizes $|V(T_w)| \leq \lceil \frac{n}{L+1} \rceil$, we have found $T_u$ and can stop.
    \item Otherwise, update $u$ to any child $w$ with subtree size $|V(T_w)| > \lceil \frac{n}{L+1} \rceil$, and go to step 2.
    Note that $|V(T_w)| \geq 2 + \lceil \frac{n}{L+1} \rceil$ from the above discussion.
\end{enumerate}

Since $|V(T_u)| > 1 + \lceil \frac{n}{L+1} \rceil$, removing $V(T_u)$ from $V$ decreases the size of $G$ by at least $\lceil \frac{n}{L+1} \rceil$.
Furthermore, each subtree pruned by $u$'s removal has size at most $\lceil \frac{n}{L+1} \rceil$ since all subtrees of children $w$ of $u$ has size at most $\lceil \frac{n}{L+1} \rceil$.

\paragraph{Running time}
Rooting the tree and computing sizes of each rooted subtree can be done in polynomial time via depth-first search and dynamic programming.
Finding the node $u$ in within the while loop can also be done in polynomial time.
Since the while loop executes at most $L \leq n$ times, the algorithm runs in polynomial time.
\end{proof}

\begin{lemma}
\label{lem:tree-r}
Suppose $\cE(G^*)$ is a tree on $n$ vertices.
There is a $r$-adaptive algorithm that uses at most $\cO(r \cdot n^{1/r})$ interventions in total.
\end{lemma}
\begin{proof}
Since the underlying causal DAG is a tree, whenever $\cE(G^*)$ is not fully oriented, the chain components are singletons and at most one tree (of size at least two).

Let $L = \lceil n^{1/r} \rceil$.
Consider the following algorithm:
\begin{enumerate}
    \item Run \cref{alg:balanced-partitioning-trees} on the tree chain component of $\cE(G^*)$ to obtain a subset of vertices $A$ of size $|A| \leq L$.
    \item Intervene on all vertices in $A$.
    \item If the essential graph is still not fully oriented and we have used less than $r-1$ rounds, go back to step 1.
    \item In the final round, we intervene on all vertices\footnote{We could have further optimized by intervening on a $G$-separating system of the remaining tree component but this does not affect the asymptotics in the worst case. For instance, if the underlying causal DAG is a single path, then the final round will still be a path and thus the $G$-separating system size is still roughly at least half the remaining nodes.} in the tree chain component (if it exists).
\end{enumerate}
In the first $r-1$ adaptive rounds, the new tree chain component must have size at most one of the pruned subtrees, i.e.\ size at most $1/(L+1)$ factor smaller than before.
After $r-1$ rounds, the largest possible size of the tree chain component is at most (via repeated applications of \cref{lem:nested-divisions}):
\[
\left\lceil \left\lceil \left\lceil  \frac{n}{L+1} \right\rceil / (L+1) \right\rceil \ldots \right\rceil
\leq \ldots
\leq \left\lceil \frac{n}{(L+1)^{r-1}} \right\rceil
\]
Since $L = \lceil n^{1/r} \rceil$, we see that $\left\lceil \frac{n}{(L+1)^{r-1}} \right\rceil \leq L$.
Therefore, by intervening on all the remaining relevant vertices in the final $r$-th round, the total number of interventions performed will be $\cO(r \cdot L) \subseteq \cO(r \cdot n^{1/r})$.
\end{proof}

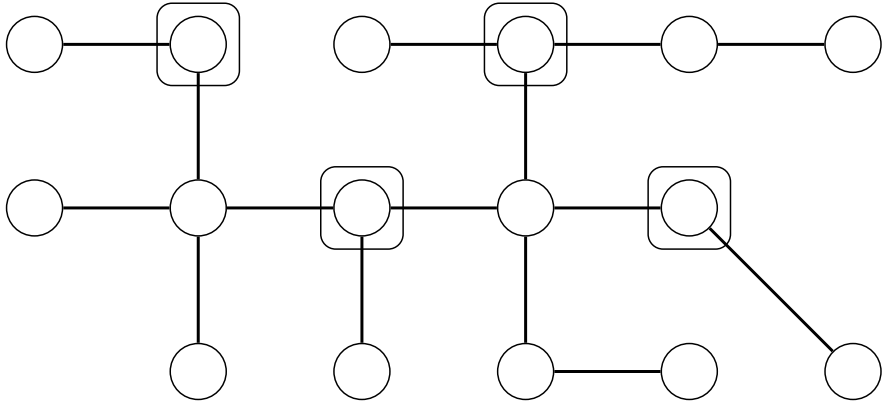
\begin{figure}[htb]
\centering
\resizebox{0.7\linewidth}{!}{%
\begin{tikzpicture}
    \node[draw, circle, minimum size=15pt, inner sep=2pt] at (0,0) (v1) {};
    \node[draw, circle, minimum size=15pt, inner sep=2pt, left=of v1] (v2) {};
    \node[draw, circle, minimum size=15pt, inner sep=2pt, above=of v2] (v3) {};
    \node[draw, circle, minimum size=15pt, inner sep=2pt, below=of v2] (v4) {};
    \node[draw, circle, minimum size=15pt, inner sep=2pt, right=of v1] (v5) {};
    \node[draw, circle, minimum size=15pt, inner sep=2pt, right=of v5] (v6) {};
    \node[draw, circle, minimum size=15pt, inner sep=2pt, above=of v5] (v7) {};
    \node[draw, circle, minimum size=15pt, inner sep=2pt, below=of v5] (v8) {};
    \node[draw, circle, minimum size=15pt, inner sep=2pt, right=of v7] (v9) {};
    \node[draw, circle, minimum size=15pt, inner sep=2pt, right=of v8] (v10) {};
    \node[draw, circle, minimum size=15pt, inner sep=2pt, left=of v7] (v11) {};
    \node[draw, circle, minimum size=15pt, inner sep=2pt, left=of v8] (v12) {};
    \node[draw, circle, minimum size=15pt, inner sep=2pt, left=of v2] (v13) {};
    \node[draw, circle, minimum size=15pt, inner sep=2pt, left=of v3] (v14) {};
    \node[draw, circle, minimum size=15pt, inner sep=2pt, right=of v9] (v15) {};
    \node[draw, circle, minimum size=15pt, inner sep=2pt, right=of v10] (v16) {};

    \draw[thick] (v1) -- (v2);
    \draw[thick] (v1) -- (v5);
    \draw[thick] (v1) -- (v12);
    \draw[thick] (v2) -- (v3);
    \draw[thick] (v2) -- (v4);
    \draw[thick] (v2) -- (v13);
    \draw[thick] (v3) -- (v14);
    \draw[thick] (v5) -- (v6);
    \draw[thick] (v5) -- (v7);
    \draw[thick] (v5) -- (v8);
    \draw[thick] (v6) -- (v16);
    \draw[thick] (v7) -- (v9);
    \draw[thick] (v7) -- (v11);
    \draw[thick] (v8) -- (v10);
    \draw[thick] (v9) -- (v15);

    \node[draw, rounded corners, fit=(v1)] {};
    \node[draw, rounded corners, fit=(v3)] {};
    \node[draw, rounded corners, fit=(v6)] {};
    \node[draw, rounded corners, fit=(v7)] {};
\end{tikzpicture}
}
\caption{Tree example on $n=16$ nodes where the boxed vertices represent a subset $A \subseteq V$ of size $|A| = \sqrt{n} = 4$. Observe that the subtrees in $G[V \setminus A]$ each are of size at most $\lceil n/(\sqrt{n} + 1) \rceil = \lceil 16/5 \rceil = 4$.}
\label{fig:tree-example}
\end{figure}

\cref{fig:tree-example} illustrates a potential partitioning example when the essential graph is a tree on $n=16$ nodes and $r = 2$.

\section{Full generality}
\label{sec:full-generality}

Building upon the intuition developed in \cref{sec:warmup}, we now show how to obtain \cref{thm:atomic-upper-bound}.
We also present a matching worst case lower bound (\cref{thm:atomic-worst-case}), and extend our results to the bounded size intervention settings (\cref{thm:bounded-upper-bound}).

Let us now describe the algorithmic idea behind our atomic intervention result.
By \cref{thm:can-remove-oriented}, we can remove all oriented edges from earlier rounds of interventions and focus on each undirected chain component.
Let $L$ be a suitably chosen value that depends on $r$, which we define later.
Since these chain components are chordal graphs, we can efficiently compute their clique tree representation for each chain component $G$.
Invoking \cref{alg:balanced-partitioning-trees} to compute a ``balanced partitioning'' on the clique tree $T_G$, we obtain $L$ maximal cliques from each chain component $G$.
Intervening on all these cliques incurs a cost of $\cO(L \cdot \nu_{1}(G^*))$ interventions, as the sum of maximal cliques in each chain component is a lower bound for the verification number (see \cref{lem:strengthened-lb}).
Performing this operation for $r-1$ rounds leaves us with chain components, each with at most $L$ maximal cliques.
In the final $r$-th round, we intervene on all the vertices in these unoriented connected components which additionally incurs a cost of at most $\cO(L \cdot \nu_{1}(G^*))$.
Therefore, the total cost in all iterations is at most $\cO(r \cdot L \cdot \nu_{1}(G^*))$.
Setting $L \approx n^{1/r}$ above gives the desired guarantees.

To formally prove our result, we first begin with following lemma which tells us that intervening on all vertices within a maximal clique of a chordal graph breaks up the chain component according to its clique tree representation.

\begin{lemma}
\label{lem:clique-tree-separation}
Let $T_G = (K, S)$ be the clique tree of a chordal graph $G = (V, E)$.
For any maximal clique $K_i$ in $T_G$, if two maximal cliques $K_1$ and $K_2$ are disjoint in $T_G[K \setminus \{K_i\}]$, then $v_1 \in V(K_1)$ and $v_2 \in V(K_2)$ are not in the same connected component of $G[V \setminus V(K_i)]$.
\end{lemma}
\begin{proof}
Since $K_1$ and $K_2$ are disjoint in $T_G[K \setminus \{K_i\}]$, the removed $K_i$ must lie on the unique path between $K_1$ and $K_2$ in the tree $T_G$.
As $V(K_i) = \cup_{\{K_i, K_j\} \in S} (V(K_i) \cap V(K_j))$, \cref{lem:prelim-lemma-5} tells us that the removal of $K_i$ will makes $v_1$ and $v_2$ disconnected in $G[V \setminus V(K_i)]$.
\end{proof}

We are now ready to prove \cref{thm:atomic-upper-bound} using \cref{alg:adaptive-search}.

\begin{algorithm}[htb]
\caption{Adaptivity-sensitive search.}
\label{alg:adaptive-search}
\begin{algorithmic}[1]
    \Statex \textbf{Input}: Essential graph $\cE(G^*)$, adaptivity param.\ $r \geq 1$.
    \Statex \textbf{Output}: A sequence of intervention sets $\cI_1, \ldots, \cI_r$ such that $\cE_{\cI_1, \ldots, \cI_r}(G^*) = G^*$.
    \State Initialize $L = \lceil n^{1/r} \rceil$.
    \For{$i = 1, \ldots, r-1$}
        \State Initialize $\cI_i \gets \emptyset$
        \For{chain comp.\ $H \in CC(\cE_{\cI_1, \ldots, \cI_{i-1}}(G^*))$}
            \If{$H$ is a clique}
                \State Set $V' \gets V(H)$.
            \Else
                \State Compute clique tree $T_H$ of $H$.
                \State Compute $L$-balanced partitioning $S$ of $T_H$ via \cref{alg:balanced-partitioning-trees}.
                \State Let $V' \gets \cup_{K_j \in S} V(K_j)$.
            \EndIf
            \State Add $V'$ to $\cI_i$.
        \EndFor
        \State Intervene on all vertices in $\cI_i$.
    \EndFor
    \State Define $\cI_r$ as all remaining relevant vertices and intervene on vertices in $\cI_r$.
    \State \Return $\cI_1, \ldots, \cI_r$
\end{algorithmic}
\end{algorithm}

To visualize the full generalization of \cref{alg:adaptive-search}, think of \cref{fig:tree-example} as the clique tree representation of a chain component of the essential graph where each node is a maximal clique.
When we intervene on a ``node'' in \cref{fig:tree-example}, we actually intervene on the entire maximally clique in an atomic fashion so that all incident edges to the clique vertices are oriented.

\atomicupperbound*
\begin{proof}
Consider \cref{alg:adaptive-search}.
Since we always intervene on all relevant vertices outside of the for loop, we are guaranteed to fully recover $G^*$.

\textbf{Number of interventions}

By \cref{lem:tree-subroutine-works}, we know that invoking \cref{alg:balanced-partitioning-trees} for any clique tree $T_H$ on maximal cliques $K_H$ of $H$ returns a set $A_H \subseteq K_H$ of at most $|A_H| \leq L$ clique nodes such that subtrees in $T_H[K_H \setminus A_H]$ have size at most $\lceil |T_H|/(L+1) \rceil$.

By \cref{lem:strengthened-lb}, 
\[
\left| \bigcup_{H \in CC(\cE_{\cI_1 \cup \ldots \cup \cI_{i-1}}(G^*))} A_H \right|
= \sum_{H \in CC(\cE_{\cI_1 \cup \ldots \cup \cI_{i-1}}(G^*))} |A_H|
\in \cO(L \cdot \nu_1(G^*))
\]
So, within the for loop, we incur at most $\cO(r \cdot L \cdot \nu_1(G^*))$ interventions.

By \cref{lem:clique-tree-separation}, we know that each iteration reduces the maximum number of maximal cliques in any chain component size by a factor of $L+1$.
After $r-1$ rounds, the largest clique tree in any chain component has at most
\[
\frac{n}{(L+1)^{r-1}}
= \frac{n}{(\lceil n^{1/r} \rceil + 1)^{r-1}}
\leq \lceil n^{1/r} \rceil
= L
\]
maximal cliques.
So, if we intervene on all remaining relevant vertices in the final $r$-th round, this incurs at most $\cO(L \cdot \nu_1(G^*))$ interventions via \cref{lem:strengthened-lb}.

Therefore, in total, we use $\cO(r \cdot L \cdot \nu_1(G^*)) \subseteq \cO(r \cdot n^{1/r} \cdot \nu_1(G^*))$ atomic interventions.

\textbf{Running time}

Throughout, executing Meek rules after performing an intervention can be done in $\cO(d \cdot m) \subseteq \cO(n \cdot m)$ time \cite{wienobst2021extendability}, where $d$ is the degeneracy of the input graph.
Now, consider an arbitrary iteration of the while loop.
There are at most $n$ chain components.
Given a chordal graph with $n$ nodes and $m$ edges, a clique tree can computed in 
$\cO(n + m)$ time \cite{blair1993introduction,galinier1995chordal}.
Given a tree with $n$ nodes, computing a $L$-balanced partitioning (\cref{alg:balanced-partitioning-trees}) takes $\cO(n)$ time using depth-first search.
So, each iteration takes $\cO(n \cdot (n + m))$ time.
Since there are $L = \lceil n^{1/r} \rceil$ iterations, the overall running time is $\cO(L \cdot n \cdot (n+m)) \subseteq \cO(n^{1+1/r} \cdot (n+m))$.
\end{proof}

\paragraph{Atomic worst case.}
Our lower bound (\cref{thm:atomic-worst-case}) generalizes the idea behind the lower bound proof in \cref{lem:path-r-2} (for the special case of $r=2$):
on a path essential graph with $\nu_1(G^*) = 1$, the adaptive adversary repeatedly hides the source node in the largest possible unoriented segment based on the current round of interventions.
See Appendix C for the full proof of \cref{thm:atomic-worst-case}.

\paragraph{Bounded size interventions.}
With bounded size interventions, each intervention is now allowed to involve $k \geq 1$ vertices, for some pre-determined upper bound $k$.
Algorithmically, we tweak Lines 7 and 11 in \cref{alg:adaptive-search} to use the labelling lemma of \cref{lem:labelling-lemma}, which incurs additional $\cO(\log k)$ multiplicative factor when comparing with $\nu_k(G^*)$.
See Appendix B for the tweaked algorithm and Appendix C for the full proof of \cref{thm:bounded-upper-bound}.

\section{Experiments}
\label{sec:experiments}

While our main contributions are theoretical, we also performed some experiments to empirically validate that \cref{alg:adaptive-search} is practical and that fewer interventions are generally needed when higher levels of adaptivity are allowed.

Using the synthetic experimental setup of \cite{squires2020active,choo2022verification,choo2023subset}, we benchmarked our adaptivity-sensitive search algorithm with varying values of $r$ together with the algorithm of \cite{choo2022verification}.
For full experimental details, implementation details, and source code, please see Appendix D.

\paragraph{Checks to avoid redundant interventions}

The current implementation of \cite{choo2022verification}'s \texttt{separator} algorithm is actually $n$-adaptive because it performs ``checks'' before performing each intervention --- if the vertices in the proposed intervention set $S$ do \emph{not} have any unoriented incident arcs, then the intervention set $S$ will be skipped.
One may think of such interventions as ``redundant'' since they do not yield any new information about the underlying causal graph.
As such, we ran two versions of their algorithm: one without checks (i.e.\ $\cO(\log n)$-adaptive) and one with checks (i.e.\ $n$-adaptive).
Note that each check corresponds to an adaptivity round because an intervention within a batch of interventions may turn out to be redundant, but we will only know this after performing a check after some of the interventions within that batch have been executed.

\paragraph{Scaling our algorithm with checks}

Since $n^{\frac{1}{\log n}} = 2$, running \cref{alg:adaptive-search} (as it is) with adaptivity parameters $r \in \Omega(\log n)$ does not make much sense.
As such, we define a checking budget $b = r - \lceil \log n \rceil$ and greedily perform up to $b$ checks whilst executing \cref{alg:adaptive-search}.
This allows \cref{alg:adaptive-search} to scale naturally for $r \in \Omega(\log n)$.

\paragraph{Empirical trends}

\cref{fig:subset} shows a subset of our results (see Appendix D for all experimental results).
As expected, we observe that higher rounds of adaptivity leads to lower number of interventions required.
When $r=n$, \cref{alg:adaptive-search} can match \cite{choo2022verification} with its full adaptivity.

\begin{figure}[htb]
\centering
\includegraphics[width=\linewidth]{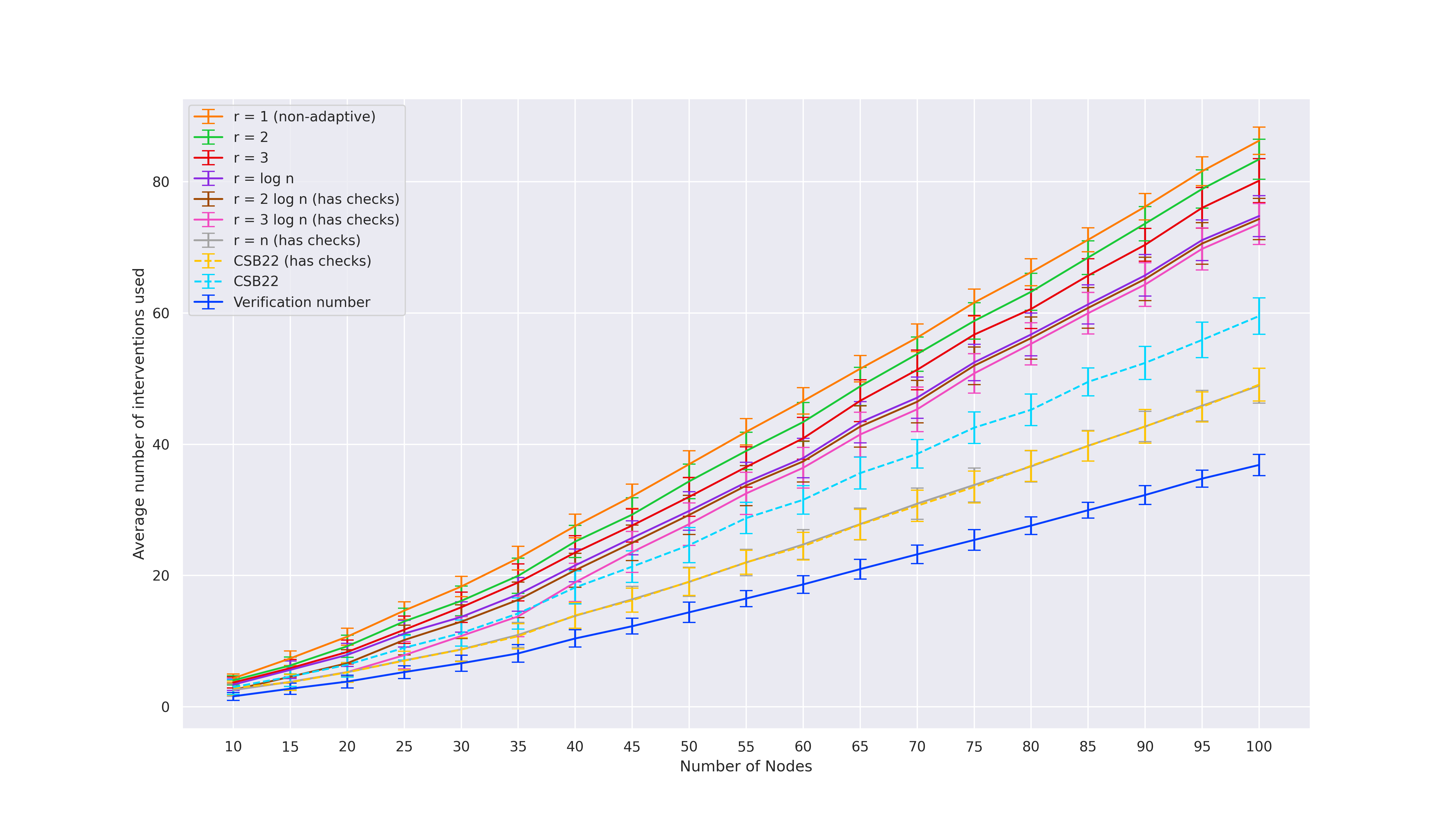}
\caption{Subset of experimental results}
\label{fig:subset}
\end{figure}

\section{Conclusion}
\label{sec:conclusion}

In our work, we define $r$-adaptivity that interpolates between non-adaptivity (for $r=1$) and full adaptivity (for $r=n$).
We provide a $r$-adaptive algorithm that achieves $\cO(\min\{r, \log n\} \cdot n^{1/\min\{r, \log n\}})$ approximation with respect to the verification number.

Let us denote $\nu^r$ as the necessary number of interventions required when allowed $r$ rounds of adaptive interventions.
Amongst $r \in \{1, 2, \ldots, n\}$, the only two values of $\nu^r$ that we currently understand are the extremes of $\nu^1$ (non-adaptive) and $\nu^n$ (full adaptivity), where $\nu^1$ is the size of a $\skel(G^*)$-separating system and $\nu^n$ corresponds to the verification number of $G^*$.
Furthermore, we also know that $\nu^r$ is non-decreasing as $r$ decreases from $n$ to $1$.
Ideally, given an input $r$, we want to compete against $\nu^r$.
Unfortunately, as we only currently understand $\nu^1$ and $\nu^n$, and $\nu^1$ could potentially even be exponentially larger than $\nu^n$ (e.g.\ when the essential graph is a path), we gave approximations in terms of $\nu^1$ in this work while it may be the case that $\nu^1$ is a weak lower bound for $r$-adaptive algorithms, especially for smaller values of $r$.
As such, understanding the correct bound for a given $r$ is an interesting open direction to pursue and with the hopes that one can design an corresponding $r$-adaptive search algorithm achieving a good approximation with respect to $\nu^r$, possibly better than \cref{alg:adaptive-search}.

Here, we studied the complexity of adaptivity for causal discovery under the standard assumptions of causal sufficiency, faithfulness, and infinite sample regime.
When these assumptions are violated, wrong causal conclusions may be drawn and possibly lead to unintended downstream consequences.
Hence, it is of great interest to remove/weaken these assumptions while maintaining theoretical guarantees.

\section*{Acknowledgements}
This research/project is supported by the National Research Foundation, Singapore under its AI Singapore Programme (AISG Award No: AISG-PhD/2021-08-013).
Part of this work was done while the authors were visiting the Simons Institute for the Theory of Computing.
We would like to thank the UAI reviewers for valuable feedback and discussions.

\bibliography{uai2023-template}

\newcommand{\etalchar}[1]{$^{#1}$}
\begin{thebibliography}{dCCCM19}

\bibitem[AAAK17]{agarwal2017learning}
Arpit Agarwal, Shivani Agarwal, Sepehr Assadi, and Sanjeev Khanna.
\newblock Learning with limited rounds of adaptivity: Coin tossing, multi-armed
  bandits, and ranking from pairwise comparisons.
\newblock In {\em Conference on Learning Theory}, pages 39--75. PMLR, 2017.

\bibitem[AMP97]{andersson1997characterization}
Steen~A. Andersson, David Madigan, and Michael~D. Perlman.
\newblock {A characterization of Markov equivalence classes for acyclic
  digraphs}.
\newblock {\em The Annals of Statistics}, 25(2):505--541, 1997.

\bibitem[ANRW15]{alon2015welfare}
Noga Alon, Noam Nisan, Ran Raz, and Omri Weinstein.
\newblock Welfare maximization with limited interaction.
\newblock In {\em 2015 IEEE 56th Annual Symposium on Foundations of Computer
  Science}, pages 1499--1512. IEEE, 2015.

\bibitem[BL22]{branzei2022query}
Simina Br{\^a}nzei and Jiawei Li.
\newblock The query complexity of local search and brouwer in rounds.
\newblock In {\em Conference on Learning Theory}, pages 5128--5145. PMLR, 2022.

\bibitem[BMW16]{braverman2016parallel}
Mark Braverman, Jieming Mao, and S~Matthew Weinberg.
\newblock Parallel algorithms for select and partition with noisy comparisons.
\newblock In {\em Proceedings of the forty-eighth annual ACM symposium on
  Theory of Computing}, pages 851--862, 2016.

\bibitem[BP93]{blair1993introduction}
Jean~RS Blair and Barry Peyton.
\newblock An introduction to chordal graphs and clique trees.
\newblock In {\em Graph theory and sparse matrix computation}, pages 1--29.
  Springer, 1993.

\bibitem[BS18]{balkanski2018adaptive}
Eric Balkanski and Yaron Singer.
\newblock The adaptive complexity of maximizing a submodular function.
\newblock In {\em Proceedings of the 50th annual ACM SIGACT symposium on theory
  of computing}, pages 1138--1151, 2018.

\bibitem[{Can}18]{canonne2018adaptivity}
{Canonne, Cl{\'e}ment L and Gur, Tom}.
\newblock {An Adaptivity Hierarchy Theorem for Property Testing}.
\newblock {\em computational complexity}, 27:671--716, 2018.

\bibitem[CBP16]{cho2016reconstructing}
Hyunghoon Cho, Bonnie Berger, and Jian Peng.
\newblock {Reconstructing Causal Biological Networks through Active Learning}.
\newblock {\em {PLoS ONE}}, 11(3):e0150611, 2016.

\bibitem[Col88]{cole1988parallel}
Richard Cole.
\newblock Parallel merge sort.
\newblock {\em SIAM Journal on Computing}, 17(4):770--785, 1988.

\bibitem[CS23]{choo2023subset}
Davin Choo and Kirankumar Shiragur.
\newblock {Subset verification and search algorithms for causal DAGs}.
\newblock In {\em International Conference on Artificial Intelligence and
  Statistics}, 2023.

\bibitem[CSB22]{choo2022verification}
Davin Choo, Kirankumar Shiragur, and Arnab Bhattacharyya.
\newblock {Verification and search algorithms for causal DAGs}.
\newblock {\em Advances in Neural Information Processing Systems}, 35, 2022.

\bibitem[dCCCM19]{de2019combining}
Luis~M. de~Campos, Andr{\'e}s Cano, Javier~G. Castellano, and Seraf{\'\i}n
  Moral.
\newblock {Combining gene expression data and prior knowledge for inferring
  gene regulatory networks via Bayesian networks using structural
  restrictions}.
\newblock {\em Statistical Applications in Genetics and Molecular Biology},
  18(3), 2019.

\bibitem[DGS84]{duris1984lower}
Pavol Duris, Zvi Galil, and Georg Schnitger.
\newblock Lower bounds on communication complexity.
\newblock In {\em Proceedings of the sixteenth annual ACM symposium on Theory
  of computing}, pages 81--91, 1984.

\bibitem[DGV08]{dean2008approximating}
Brian~C Dean, Michel~X Goemans, and Jan Vondr{\'a}k.
\newblock Approximating the stochastic knapsack problem: The benefit of
  adaptivity.
\newblock {\em Mathematics of Operations Research}, 33(4):945--964, 2008.

\bibitem[Ebe07]{eberhardt2007causation}
Frederick Eberhardt.
\newblock {Causation and Intervention}.
\newblock {\em Unpublished doctoral dissertation, Carnegie Mellon University},
  page~93, 2007.

\bibitem[Ebe10]{eberhardt2010causal}
Frederick Eberhardt.
\newblock {Causal Discovery as a Game}.
\newblock In {\em Causality: Objectives and Assessment}, pages 87--96. PMLR,
  2010.

\bibitem[EGS05]{eberhardt2005number}
Frederick Eberhardt, Clark Glymour, and Richard Scheines.
\newblock {On the number of experiments sufficient and in the worst case
  necessary to identify all causal relations among N variables}.
\newblock In {\em Proceedings of the Twenty-First Conference on Uncertainty in
  Artificial Intelligence}, pages 178--184, 2005.

\bibitem[EGS06]{eberhardt2006n}
Frederick Eberhardt, Clark Glymour, and Richard Scheines.
\newblock {N-1 Experiments Suffice to Determine the Causal Relations Among N
  Variables}.
\newblock In {\em Innovations in machine learning}, pages 97--112. Springer,
  2006.

\bibitem[EGS12]{eberhardt2012number}
Frederick Eberhardt, Clark Glymour, and Richard Scheines.
\newblock {On the Number of Experiments Sufficient and in the Worst Case
  Necessary to Identify All Causal Relations Among N Variables}.
\newblock {\em arXiv preprint arXiv:1207.1389}, 2012.

\bibitem[ES07]{eberhardt2007interventions}
Frederick Eberhardt and Richard Scheines.
\newblock {Interventions and Causal Inference}.
\newblock {\em Philosophy of science}, 74(5):981--995, 2007.

\bibitem[FG65]{fulkerson1965incidence}
Delbert Fulkerson and Oliver Gross.
\newblock Incidence matrices and interval graphs.
\newblock {\em Pacific journal of mathematics}, 15(3):835--855, 1965.

\bibitem[Gav72]{gavril1972algorithms}
F{\u{a}}nic{\u{a}} Gavril.
\newblock Algorithms for minimum coloring, maximum clique, minimum covering by
  cliques, and maximum independent set of a chordal graph.
\newblock {\em SIAM Journal on Computing}, 1(2):180--187, 1972.

\bibitem[GHP95]{galinier1995chordal}
Philippe Galinier, Michel Habib, and Christophe Paul.
\newblock Chordal graphs and their clique graphs.
\newblock {\em WG}, 95:358--371, 1995.

\bibitem[GKP94]{graham1994concrete}
Ronald~L Graham, Donald~E Knuth, and Oren Patashnik.
\newblock Concrete mathematics: A foundation for computer science, 1994.

\bibitem[GKS{\etalchar{+}}19]{greenewald2019sample}
Kristjan Greenewald, Dmitriy Katz, Karthikeyan Shanmugam, Sara Magliacane,
  Murat Kocaoglu, Enric Boix-Adser\`{a}, and Guy Bresler.
\newblock {Sample Efficient Active Learning of Causal Trees}.
\newblock {\em Advances in Neural Information Processing Systems}, 32, 2019.

\bibitem[GNS16]{gupta2016algorithms}
Anupam Gupta, Viswanath Nagarajan, and Sahil Singla.
\newblock Algorithms and adaptivity gaps for stochastic probing.
\newblock In {\em Proceedings of the twenty-seventh annual ACM-SIAM symposium
  on Discrete algorithms}, pages 1731--1747. SIAM, 2016.

\bibitem[GNS17]{gupta2017adaptivity}
Anupam Gupta, Viswanath Nagarajan, and Sahil Singla.
\newblock Adaptivity gaps for stochastic probing: Submodular and xos functions.
\newblock In {\em Proceedings of the Twenty-Eighth Annual ACM-SIAM Symposium on
  Discrete Algorithms}, pages 1688--1702. SIAM, 2017.

\bibitem[GV06]{goemans2006stochastic}
Michel Goemans and Jan Vondr{\'a}k.
\newblock Stochastic covering and adaptivity.
\newblock In {\em LATIN 2006: Theoretical Informatics: 7th Latin American
  Symposium, Valdivia, Chile, March 20-24, 2006. Proceedings 7}, pages
  532--543. Springer, 2006.

\bibitem[HB12]{hauser2012characterization}
Alain Hauser and Peter B\"{u}hlmann.
\newblock {Characterization and greedy learning of interventional Markov
  equivalence classes of directed acyclic graphs}.
\newblock {\em The Journal of Machine Learning Research}, 13(1):2409--2464,
  2012.

\bibitem[HB14]{hauser2014two}
Alain Hauser and Peter B\"{u}hlmann.
\newblock {Two Optimal Strategies for Active Learning of Causal Models from
  Interventions}.
\newblock {\em International Journal of Approximate Reasoning}, 55(4):926--939,
  2014.

\bibitem[HEH13]{hyttinen2013experiment}
Antti Hyttinen, Frederick Eberhardt, and Patrik~O. Hoyer.
\newblock {Experiment Selection for Causal Discovery}.
\newblock {\em Journal of Machine Learning Research}, 14:3041--3071, 2013.

\bibitem[HLV14]{hu2014randomized}
Huining Hu, Zhentao Li, and Adrian Vetta.
\newblock {Randomized Experimental Design for Causal Graph Discovery}.
\newblock {\em Advances in Neural Information Processing Systems}, 27, 2014.

\bibitem[HNC09]{haupt2009adaptive}
Jarvis Haupt, Robert Nowak, and Rui Castro.
\newblock {Adaptive Sensing for Sparse Recovery}.
\newblock In {\em 2009 IEEE 13th Digital Signal Processing Workshop and 5th
  IEEE Signal Processing Education Workshop}, pages 702--707. IEEE, 2009.

\bibitem[Hoo90]{hoover1990logic}
Kevin~D Hoover.
\newblock {The logic of causal inference: Econometrics and the Conditional
  Analysis of Causation}.
\newblock {\em Economics \& Philosophy}, 6(2):207--234, 1990.

\bibitem[IPW11]{indyk2011power}
Piotr Indyk, Eric Price, and David~P Woodruff.
\newblock {On the Power of Adaptivity in Sparse Recovery}.
\newblock In {\em 2011 IEEE 52nd Annual Symposium on Foundations of Computer
  Science}, pages 285--294. IEEE, 2011.

\bibitem[KDV17]{kocaoglu2017cost}
Murat Kocaoglu, Alex Dimakis, and Sriram Vishwanath.
\newblock {Cost-Optimal Learning of Causal Graphs}.
\newblock In {\em International Conference on Machine Learning}, pages
  1875--1884. PMLR, 2017.

\bibitem[KF09]{koller2009probabilistic}
Daphne Koller and Nir Friedman.
\newblock {\em Probabilistic graphical models: principles and techniques}.
\newblock MIT press, 2009.

\bibitem[KWJ{\etalchar{+}}04]{king2004functional}
Ross~D. King, Kenneth~E. Whelan, Ffion~M. Jones, Philip G.~K. Reiser,
  Christopher~H. Bryant, Stephen~H. Muggleton, Douglas~B. Kell, and Stephen~G.
  Oliver.
\newblock {Functional genomic hypothesis generation and experimentation by a
  robot scientist}.
\newblock {\em Nature}, 427(6971):247--252, 2004.

\bibitem[Leu84]{leung1984fast}
Joseph Y-T Leung.
\newblock Fast algorithms for generating all maximal independent sets of
  interval, circular-arc and chordal graphs.
\newblock {\em Journal of Algorithms}, 5(1):22--35, 1984.

\bibitem[Mee95]{meek1995}
Christopher Meek.
\newblock {Causal Inference and Causal Explanation with Background Knowledge}.
\newblock In {\em Proceedings of the Eleventh Conference on Uncertainty in
  Artificial Intelligence}, UAI'95, page 403–410, San Francisco, CA, USA,
  1995. Morgan Kaufmann Publishers Inc.

\bibitem[MSW08]{malioutov2008compressed}
Dmitry~M Malioutov, Sujay Sanghavi, and Alan~S Willsky.
\newblock Compressed sensing with sequential observations.
\newblock In {\em 2008 IEEE International Conference on Acoustics, Speech and
  Signal Processing}, pages 3357--3360. IEEE, 2008.

\bibitem[NW91]{nisan1991rounds}
Noam Nisan and Avi Widgerson.
\newblock Rounds in communication complexity revisited.
\newblock In {\em Proceedings of the twenty-third annual ACM symposium on
  Theory of computing}, pages 419--429, 1991.

\bibitem[Pea09]{pearl2009causality}
Judea Pearl.
\newblock {\em {Causality: Models, Reasoning and Inference}}.
\newblock Cambridge University Press, USA, 2nd edition, 2009.

\bibitem[POS{\etalchar{+}}18]{pingault2018using}
Jean-Baptiste Pingault, Paul~F O'reilly, Tabea Schoeler, George~B Ploubidis,
  Fr{\"u}hling Rijsdijk, and Frank Dudbridge.
\newblock {Using genetic data to strengthen causal inference in observational
  research}.
\newblock {\em Nature Reviews Genetics}, 19(9):566--580, 2018.

\bibitem[PS82]{papadimitriou1982communication}
Christos~H Papadimitriou and Michael Sipser.
\newblock Communication complexity.
\newblock In {\em Proceedings of the fourteenth annual ACM symposium on Theory
  of computing}, pages 196--200, 1982.

\bibitem[Rei56]{reichenbach1956direction}
Hans Reichenbach.
\newblock {\em {The Direction of Time}}, volume~65.
\newblock University of California Press, 1956.

\bibitem[RHT{\etalchar{+}}17]{rotmensch2017learning}
Maya Rotmensch, Yoni Halpern, Abdulhakim Tlimat, Steven Horng, and David
  Sontag.
\newblock {Learning a Health Knowledge Graph from Electronic Medical Records}.
\newblock {\em Scientific reports}, 7(1):1--11, 2017.

\bibitem[RS06]{raskhodnikova2006note}
Sofya Raskhodnikova and Adam Smith.
\newblock {A Note on Adaptivity in Testing Properties of Bounded Degree
  Graphs}.
\newblock In {\em Electronic Colloquium on Computational Complexity (ECCC)}.
  Citeseer, 2006.

\bibitem[RTL76]{rose1976algorithmic}
Donald~J. Rose, R.~Endre Tarjan, and George~S. Lueker.
\newblock {Algorithmic Aspects of Vertex Elimination on Graphs}.
\newblock {\em SIAM Journal on Computing}, 5(2):266--283, 1976.

\bibitem[RW06]{rubin2006estimating}
Donald~B Rubin and Richard~P Waterman.
\newblock {Estimating the Causal Effects of Marketing Interventions Using
  Propensity Score Methodology}.
\newblock {\em Statistical Science}, pages 206--222, 2006.

\bibitem[SC17]{sverchkov2017review}
Yuriy Sverchkov and Mark Craven.
\newblock {A review of active learning approaches to experimental design for
  uncovering biological networks}.
\newblock {\em PLoS computational biology}, 13(6):e1005466, 2017.

\bibitem[Sca18]{scarlett2018noisy}
Jonathan Scarlett.
\newblock Noisy adaptive group testing: Bounds and algorithms.
\newblock {\em IEEE Transactions on Information Theory}, 65(6):3646--3661,
  2018.

\bibitem[SKDV15]{shanmugam2015learning}
Karthikeyan Shanmugam, Murat Kocaoglu, Alexandros~G. Dimakis, and Sriram
  Vishwanath.
\newblock {Learning Causal Graphs with Small Interventions}.
\newblock {\em Advances in Neural Information Processing Systems}, 28, 2015.

\bibitem[SMG{\etalchar{+}}20]{squires2020active}
Chandler Squires, Sara Magliacane, Kristjan Greenewald, Dmitriy Katz, Murat
  Kocaoglu, and Karthikeyan Shanmugam.
\newblock {Active Structure Learning of Causal DAGs via Directed Clique Trees}.
\newblock {\em Advances in Neural Information Processing Systems},
  33:21500--21511, 2020.

\bibitem[Tia16]{tian2016bayesian}
Tianhai Tian.
\newblock {Bayesian Computation Methods for Inferring Regulatory Network Models
  Using Biomedical Data}.
\newblock {\em Translational Biomedical Informatics: A Precision Medicine
  Perspective}, pages 289--307, 2016.

\bibitem[Val75]{valiant1975parallelism}
Leslie~G Valiant.
\newblock Parallelism in comparison problems.
\newblock {\em SIAM Journal on Computing}, 4(3):348--355, 1975.

\bibitem[VP90]{verma1990}
Thomas Verma and Judea Pearl.
\newblock {Equivalence and Synthesis of Causal Models}.
\newblock In {\em Proceedings of the Sixth Annual Conference on Uncertainty in
  Artificial Intelligence}, UAI '90, page 255–270, USA, 1990. Elsevier
  Science Inc.

\bibitem[WBL21a]{wienobst2021extendability}
Marcel Wien{\"o}bst, Max Bannach, and Maciej Li{\'s}kiewicz.
\newblock {Extendability of Causal Graphical Models: Algorithms and
  Computational Complexity}.
\newblock In {\em Uncertainty in Artificial Intelligence}, pages 1248--1257.
  PMLR, 2021.

\bibitem[WBL21b]{pmlr-v161-wienobst21a}
Marcel Wien\"{o}bst, Max Bannach, and Maciej Li\'{s}kiewicz.
\newblock {Extendability of causal graphical models: Algorithms and
  computational complexity}.
\newblock In Cassio de~Campos and Marloes~H. Maathuis, editors, {\em
  Proceedings of the Thirty-Seventh Conference on Uncertainty in Artificial
  Intelligence}, volume 161 of {\em Proceedings of Machine Learning Research},
  pages 1248--1257. PMLR, 27--30 Jul 2021.

\bibitem[Woo05]{woodward2005making}
James Woodward.
\newblock {\em {Making Things Happen: A Theory of Causal Explanation}}.
\newblock Oxford University Press, 2005.

\end{thebibliography}
\bibliographystyle{alpha}

\newpage
\appendix

\section{Meek rules}
\label{sec:appendix-meek-rules}

Meek rules are a set of 4 edge orientation rules that are sound and complete with respect to any given set of arcs that has a consistent DAG extension \cite{meek1995}.
Given any edge orientation information, one can always repeatedly apply Meek rules till a fixed point to maximize the number of oriented arcs.

\begin{definition}[Consistent extension]
A set of arcs is said to have a \emph{consistent DAG extension} $\pi$ for a graph $G$ if there exists a permutation on the vertices such that (i) every edge $\{u,v\}$ in $G$ is oriented $u \to v$ whenever $\pi(u) < \pi(v)$, (ii) there is no directed cycle, (iii) all the given arcs are present.
\end{definition}

\begin{figure}[htb]
\centering
\resizebox{\linewidth}{!}{%
\begin{tikzpicture}
%
%
\node[draw, circle, inner sep=2pt] at (0,0) (R1a-before) {\small $a$};
\node[draw, circle, inner sep=2pt, right=of R1a-before] (R1b-before) {\small $b$};
\node[draw, circle, inner sep=2pt, above=of R1a-before](R1c-before) {\small $c$};
\draw[thick, -stealth] (R1c-before) -- (R1a-before);
\draw[thick] (R1a-before) -- (R1b-before);

\node[draw, circle, inner sep=2pt] at (3,0) (R1a-after) {\small $a$};
\node[draw, circle, inner sep=2pt, right=of R1a-after] (R1b-after) {\small $b$};
\node[draw, circle, inner sep=2pt, above=of R1a-after](R1c-after) {\small $c$};
\draw[thick, -stealth] (R1c-after) -- (R1a-after);
\draw[thick, -stealth] (R1a-after) -- (R1b-after);

\node[single arrow, draw, minimum height=2em, single arrow head extend=1ex, inner sep=2pt] at (2.2,0.75) (R1arrow) {};
\node[above=5pt of R1arrow] {\footnotesize R1};

%
%
\node[draw, circle, inner sep=2pt] at (6,0) (R2a-before) {\small $a$};
\node[draw, circle, inner sep=2pt, right=of R2a-before] (R2b-before) {\small $b$};
\node[draw, circle, inner sep=2pt, above=of R2a-before](R2c-before) {\small $c$};
\draw[thick, -stealth] (R2a-before) -- (R2c-before);
\draw[thick, -stealth] (R2c-before) -- (R2b-before);
\draw[thick] (R2a-before) -- (R2b-before);

\node[draw, circle, inner sep=2pt] at (9,0) (R2a-after) {\small $a$};
\node[draw, circle, inner sep=2pt, right=of R2a-after] (R2b-after) {\small $b$};
\node[draw, circle, inner sep=2pt, above=of R2a-after](R2c-after) {\small $c$};
\draw[thick, -stealth] (R2a-after) -- (R2c-after);
\draw[thick, -stealth] (R2c-after) -- (R2b-after);
\draw[thick, -stealth] (R2a-after) -- (R2b-after);

\node[single arrow, draw, minimum height=2em, single arrow head extend=1ex, inner sep=2pt] at (8.2,0.75) (R2arrow) {};
\node[above=5pt of R2arrow] {\footnotesize R2};

%
%
\node[draw, circle, inner sep=2pt] at (12,0) (R3d-before) {\small $d$};
\node[draw, circle, inner sep=2pt, above=of R3d-before](R3a-before) {\small $a$};
\node[draw, circle, inner sep=2pt, right=of R3a-before] (R3c-before) {\small $c$};
\node[draw, circle, inner sep=2pt, right=of R3d-before](R3b-before) {\small $b$};
\draw[thick, -stealth] (R3c-before) -- (R3b-before);
\draw[thick, -stealth] (R3d-before) -- (R3b-before);
\draw[thick] (R3c-before) -- (R3a-before) -- (R3d-before);
\draw[thick] (R3a-before) -- (R3b-before);

\node[draw, circle, inner sep=2pt] at (15,0) (R3d-after) {\small $d$};
\node[draw, circle, inner sep=2pt, above=of R3d-after](R3a-after) {\small $a$};
\node[draw, circle, inner sep=2pt, right=of R3a-after] (R3c-after) {\small $c$};
\node[draw, circle, inner sep=2pt, right=of R3d-after](R3b-after) {\small $b$};
\draw[thick, -stealth] (R3c-after) -- (R3b-after);
\draw[thick, -stealth] (R3d-after) -- (R3b-after);
\draw[thick] (R3c-after) -- (R3a-after) -- (R3d-after);
\draw[thick, -stealth] (R3a-after) -- (R3b-after);

\node[single arrow, draw, minimum height=2em, single arrow head extend=1ex, inner sep=2pt] at (14.2,0.75) (R3arrow) {};
\node[above=5pt of R3arrow] {\footnotesize R3};

%
%
\node[draw, circle, inner sep=2pt] at (18,0) (R4a-before) {\small $a$};
\node[draw, circle, inner sep=2pt, above=of R4a-before](R4d-before) {\small $d$};
\node[draw, circle, inner sep=2pt, right=of R4d-before] (R4c-before) {\small $c$};
\node[draw, circle, inner sep=2pt, right=of R4a-before](R4b-before) {\small $b$};
\draw[thick, -stealth] (R4d-before) -- (R4c-before);
\draw[thick, -stealth] (R4c-before) -- (R4b-before);
\draw[thick] (R4d-before) -- (R4a-before) -- (R4c-before);
\draw[thick] (R4a-before) -- (R4b-before);

\node[draw, circle, inner sep=2pt] at (21,0) (R4a-after) {\small $a$};
\node[draw, circle, inner sep=2pt, above=of R4a-after](R4d-after) {\small $d$};
\node[draw, circle, inner sep=2pt, right=of R4d-after] (R4c-after) {\small $c$};
\node[draw, circle, inner sep=2pt, right=of R4a-after](R4b-after) {\small $b$};
\draw[thick, -stealth] (R4d-after) -- (R4c-after);
\draw[thick, -stealth] (R4c-after) -- (R4b-after);
\draw[thick] (R4d-after) -- (R4a-after) -- (R4c-after);
\draw[thick, -stealth] (R4a-after) -- (R4b-after);

\node[single arrow, draw, minimum height=2em, single arrow head extend=1ex, inner sep=2pt] at (20.2,0.75) (R4arrow) {};
\node[above=5pt of R4arrow] {\footnotesize R4};

\draw[thick] (5.25,1.75) -- (5.25,-0.25);
\draw[thick] (11.25,1.75) -- (11.25,-0.25);
\draw[thick] (17.25,1.75) -- (17.25,-0.25);
\end{tikzpicture}
}
\caption{An illustration of the four Meek rules}
\label{fig:meek-rules}
\end{figure}
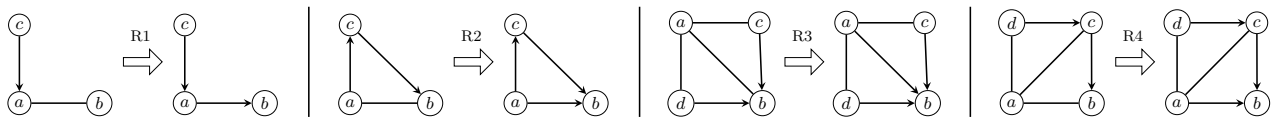

\begin{definition}[The four Meek rules \cite{meek1995}, see \cref{fig:meek-rules} for an illustration]
\hspace{0pt}
\begin{description}
    \item [R1] Edge $\{a,b\} \in E \setminus A$ is oriented as $a \to b$ if $\exists$ $c \in V$ such that $c \to a$ and $c \not\sim b$.
    \item [R2] Edge $\{a,b\} \in E \setminus A$ is oriented as $a \to b$ if $\exists$ $c \in V$ such that $a \to c \to b$.
    \item [R3] Edge $\{a,b\} \in E \setminus A$ is oriented as $a \to b$ if $\exists$ $c,d \in V$ such that $d \sim a \sim c$, $d \to b \gets c$, and $c \not\sim d$.
    \item [R4] Edge $\{a,b\} \in E \setminus A$ is oriented as $a \to b$ if $\exists$ $c,d \in V$ such that $d \sim a \sim c$, $d \to c \to b$, and $b \not\sim d$.
\end{description}
\end{definition}

There exists an algorithm \cite[Algorithm 2]{pmlr-v161-wienobst21a} that runs in $\cO(d \cdot |E|)$ time and computes the closure under Meek rules, where $d$ is the degeneracy of the graph skeleton\footnote{A $d$-degenerate graph is an undirected graph in which every subgraph has a vertex of degree at most $d$. Note that the degeneracy of a graph is typically smaller than the maximum degree of the graph.}.

\section{Deferred details}
\label{sec:appendix-deferred-details}

\subsection{Basic results}
\label{sec:appendix-basic-results}

\begin{lemma}[Equation 3.10 of \cite{graham1994concrete}]
\label{lem:eq310}
Let $f(x)$ be any continuous, monotonically increasing function with the property that $x$ is an integer if $f(x)$ is an integer.
Then, $\lceil f(x) \rceil = \lceil f(\lceil x \rceil) \rceil$.
\end{lemma}

\nesteddivisions*
\begin{proof}
Apply \cref{lem:eq310} with the function as $f(x) = x/n$ on input as $x/m$.
\end{proof}

\begin{lemma}
\label{lem:lowerboundineq}
For $r \geq 2$, we have
$
\frac{r-1}{2} \cdot \left( \frac{2}{r} \right)^{\frac{1}{r-1}}
\geq \frac{r}{4}
$.
\end{lemma}
\begin{proof}
Multiplying the left-hand side by $4/r$, we get
\begin{align*}
(r-1) \cdot \left( \frac{2}{r} \right)^{1 + \frac{1}{r-1}}
& \geq (r-1) \cdot \left( \frac{2}{r} \right) && \text{Since $r > 1$}\\
& \geq 1 && \text{Since $r \geq 2$}
\end{align*}
Thus, the inequality holds.
\end{proof}

\begin{lemma}[Theorem 12 of \cite{choo2022verification}]
\label{lem:bounded-size-lb}
For any causal DAG $G^*$, we have $\nu_k(G^*) \geq \lceil \frac{\nu_1(G^*)}{k} \rceil$.
\end{lemma}

\subsection{Algorithm for bounded size interventions}

\begin{algorithm}[H]
\caption{Adaptivity-sensitive search.}
\label{alg:adaptive-search-bounded}
\begin{algorithmic}[1]
    \Statex \textbf{Input}: Essential graph $\cE(G^*)$, adaptivity round parameter $r \geq 1$, intervention size upper bound $k \geq 1$.
    \Statex \textbf{Output}: A sequence of intervention sets $\cI_1, \ldots, \cI_r$ such that $\cE_{\cI_1, \ldots, \cI_r}(G^*) = G^*$ and $|I| \leq k$ for any intervention in $I \in \cI_i$ in intervention set $\cI_i$, $1 \leq i \leq r$.
    \State Initialize $L = \lceil n^{1/r} \rceil$.
    \For{$i = 1, \ldots, r-1$}
        \State Initialize $\cI_i \gets \emptyset$
        \For{chain comp.\ $H \in CC(\cE_{\cI_1, \ldots, \cI_{i-1}}(G^*))$}
            \If{$H$ is a clique}
                \State Set $V' \gets V(H)$.
            \Else
                \State Compute clique tree $T_H$ of $H$.
                \State Compute $L$-balanced partitioning $S$ of $T_H$ via Algorithm 1.
                \State Let $V' \gets \cup_{K_j \in S} V(K_j)$.
            \EndIf
            \State Add output of \cref{alg:intervention-subroutine} on $V'$ to $\cI$.
        \EndFor
        \State Intervene on all interventions in $\cI_i$.
    \EndFor
    \State Define $\cI_r$ as output of \cref{alg:intervention-subroutine} on remaining relevant vertices and intervene on all interventions in $\cI_r$.
    \State \Return $\cI_1, \ldots, \cI_r$
\end{algorithmic}
\end{algorithm}

\begin{algorithm}[H]
\caption{Intervention subroutine.}
\label{alg:intervention-subroutine}
\begin{algorithmic}[1]
    \Statex \textbf{Input}: Set of vertices $A$, size upper bound $k \geq 1$.
    \Statex \textbf{Output}: A $k$-separating system $B \subseteq 2^A$.
    \If{$k = 1$}
        \State Set $B \gets A$.
    \Else
        \State Define $k' = \min\{k, |A|/2\}$, $a = \lceil |A|/k' \rceil \geq 2$, and $\ell = \lceil \log_a n \rceil$.
        \State Compute labelling scheme of \cite[Lemma 1]{shanmugam2015learning} on $A$ with $(|A|, k', a)$.
        \State Set $B \gets \{S_{x,y}\}_{x \in [\ell], y \in [a]}$, where $S_{x,y} \subseteq A$ is the subset of vertices whose $x^{th}$ letter in the label is $y$.
    \EndIf
    \State \Return $B$
\end{algorithmic}
\end{algorithm}

\section{Deferred proofs}
\label{sec:appendix-proofs}

\atomicworstcase*
\begin{proof}
Without loss of generality, we may assume $r \leq \log n$ and prove a lower bound of $\Omega(r \cdot n^{1/r} \cdot \nu_1(G^*))$.

Consider the case where the essential graph is a path on $n$ nodes and the adversary can adaptively choose the source node as long as it is consistent with the arc directions revealed thus far.
On a path essential graph, $\nu(G^*) = 1$.

Suppose $r = 1$.
Then, by Theorem 13, we need to intervene on a $G$-separating system, which has size $\Omega(n)$.
The claim follows since $\nu(G^*) = 1$.

Now, suppose $r \geq 2$.
If currently have length $\ell$ segment and $k$ interventions are performed, then there must be some segment of length at least $\ell/(k+1)$.
Recurse on that.
If the final round has length $\ell$ segment, need at least $\ell/2$ interventions because $G$-separating system on a segment of length $\ell$ has size at least $\ell/2$.

Suppose the algorithm intervenes on $k_i$ vertices on the $i$-th round, for $1 \leq i \leq r$.
where $k_i \geq 1$, so $k_i + 1 \leq 2 k_i$ and so $1/(k_i + 1) \geq 1/(2 k_i)$.

Then, from the above discussion,
\begin{align*}
k_r
& \geq \frac{1}{2} \cdot n \cdot \frac{1}{k_1 + 1} \cdot \frac{1}{k_2 + 1} \cdot \ldots \cdot \frac{1}{k_{r-1} + 1}\\
& \geq \frac{1}{2^r} \cdot \frac{n}{k_1 \cdot k_2 \ldots \cdot k_{r-1}}
\end{align*}

So, the number of overall interventions used is
\begin{align*}
&\; k_1 + \ldots + k_r\\
\geq &\; k_1 + \ldots + k_{r-1} + \frac{1}{2^r} \cdot \frac{n}{k_1 \cdot k_2 \ldots \cdot k_{r-1}}\\
\geq &\; (r-1) \cdot \left( \prod_{i=1}^{r-1} k_i \right)^{\frac{1}{r-1}}
+ \frac{1}{2^r} \cdot \frac{n}{k_1 \cdot k_2 \ldots \cdot k_{r-1}}
\end{align*}
where the last inequality is the AM-GM inequality.

Let $x = k_1 \cdot k_2 \ldots \cdot k_{r-1}$.
Then,
\[
\sum_{i=1}^r k_i
= k_1 + \ldots + k_r
\geq (r-1) \cdot x^{\frac{1}{r-1}} + \frac{1}{2^r} \cdot \frac{n}{x}
\]

\textbf{Case 1}: $\frac{1}{2^r} \cdot \frac{n}{x} \geq \frac{r}{4} \cdot n^{\frac{1}{r}}$

Then,
\[
\sum_{i=1}^r k_i
\geq \frac{1}{2^r} \cdot \frac{n}{x}
\geq \frac{r}{4} \cdot n^{\frac{1}{r}}
\in \Omega(r \cdot n^{\frac{1}{r}})
\]
Thus, the claim holds as $\nu(G^*) = 1$.

\textbf{Case 2}: $\frac{1}{2^r} \cdot \frac{n}{x} < \frac{r}{4} \cdot n^{\frac{1}{r}}$

Then,
\[
x
> \frac{4 \cdot n^{1 - 1/r}}{2^r \cdot r}
= \frac{2 \cdot n^{\frac{r-1}{r}}}{2^{r-1} \cdot r}
\]
and \cref{lem:lowerboundineq} in \cref{sec:appendix-basic-results} tells us that
\begin{align*}
(r-1) \cdot x^{\frac{1}{r-1}}
& > n^{\frac{1}{r}} \cdot \frac{r-1}{2} \cdot \left( \frac{2}{r} \right)^{\frac{1}{r-1}}\\
& \geq n^{\frac{1}{r}} \cdot \frac{r}{4} && \text{For $r \geq 2$}
\end{align*}
So,
\[
\sum_{i=1}^r k_i
\geq (r-1) \cdot x^{\frac{1}{r-1}}
\geq \frac{r}{4} \cdot n^{\frac{1}{r}}
\in \Omega(r \cdot n^{\frac{1}{r}})
\]
Thus, the claim holds as $\nu(G^*) = 1$.
\end{proof}

\boundedupperbound*
\begin{proof}
We invoke \cref{alg:adaptive-search-bounded} with $k > 1$.

\textbf{Number of interventions}

The high level proof approach for is exactly the same as the proof of Theorem 1, except for how to compute intervention sets from the maximal clique vertices (obtained by ``balanced partitioning'' in the first $r-1$ rounds, within the while loop) and the from the remaining relevant vertices (in the final $r$-th round, outside the while loop).

In each iteration of the while-loop, we intervene on at most $L$ cliques for each connected component.
To orient the edges incident to these cliques we use the labelling scheme of Lemma 14 via \cref{alg:intervention-subroutine}.
So, the number of bounded size interventions we perform per round is
\[
\cO \left( L \cdot \log k \cdot \frac{\nu_{1}(G^*)}{k} \right)
\]
By \cref{lem:bounded-size-lb}, we know that $\nu_k(G^*) \geq \lceil \frac{\nu_1(G^*)}{k} \rceil$.
So, we can re-express the above bound as $\cO \left( L \cdot \log k \cdot \nu_{k}(G^*) \right)$.
Similarly, we use $\cO \left( L \cdot \log k \cdot \nu_{k}(G^*) \right)$ bounded size interventions in the final round.
Thus, over all $r$ adaptive rounds, we use a total of
\[
\cO \left( r \cdot L \cdot \log k \cdot \nu_{k}(G^*) \right)
\]
bounded size interventions.
Substituting $L = \lceil n^{1/r} \rceil$ yields our desired bound.

\textbf{Running time}

\cref{alg:adaptive-search-bounded} only differs from Algorithm 2 by invoking \cref{alg:intervention-subroutine}, which runs in polynomial time (see Lemma 14).
Thus, \cref{alg:adaptive-search-bounded} runs in polynomial time.
\end{proof}

\section{Experiments}
\label{sec:appendix-experiments}

The experiments are conducted on an Ubuntu server with two AMD EPYC 7532 CPU and 256GB DDR4 RAM.
Our code and entire experimental setup is available at\\
\url{https://github.com/cxjdavin/adaptivity-complexity-for-causal-graph-discovery}.

\subsection{Implementation details}

\paragraph{Checks to avoid redundant interventions}

The current implementation of \cite{choo2022verification}'s \texttt{separator} algorithm is actually $n$-adaptive because it performs ``checks'' before performing each intervention --- if the vertices in the proposed intervention set $S$ do \emph{not} have any unoriented incident arcs, then the intervention set $S$ will be skipped.
One may think of such interventions as ``redundant'' since they do not yield any new information about the underlying causal graph.
As such, we ran two versions of their algorithm: one without checks (i.e.\ $\cO(\log n)$-adaptive) and one with checks (i.e.\ $n$-adaptive).
Note that each check corresponds to an adaptivity round because an intervention within a batch of interventions may turn out to be redundant, but we will only know this after performing a check after some of the interventions within that batch have been executed.

\paragraph{Scaling our algorithm with checks}

Since $n^{\frac{1}{\log n}} = 2$, running Algorithm 2 (as it is) with adaptivity parameters $r \in \Omega(\log n)$ does not make much sense.
As such, we define a checking budget $b = r - \lceil \log n \rceil$ and greedily perform up to $b$ checks whilst executing Algorithm 2.
This allows Algorithm 2 to scale naturally for $r \in \Omega(\log n)$.

\paragraph{Non-adaptive intervention round}

For the final round of interventions, let $V'$ be the set of remaining relevant vertices.
From our algorithm, we know that $|V'| \leq L$ but we may even intervene on less vertices in the final round.
By \cite{kocaoglu2017cost}, we only need to intervene on a graph-separating system of the subgraph $G[V']$.
For atomic interventions, this exactly correspond to the minimum vertex cover of $V'$.
To obtain this, we first compute the maximum independent set $S$ of $V'$ (which can be computed efficiently on chordal graphs \cite{gavril1972algorithms,leung1984fast}), then only intervene on $V' \setminus S$.

\paragraph{Optimization before final round}

Note that we can always compute the intervention set $F \subseteq V$ which we \emph{would} have intervened if $r=1$.
At any point in time of the algorithm, if $F$ involves less vertices than the number of vertices required from the $L$-partitioning, then we simply treat the current adaptivity round as the final round, choose to intervene on $F$ and use any remaining adaptive budget for performing checks.

\subsection{Synthetic graphs}

We use synthetic moral randomly generated graphs from earlier prior works \cite{choo2022verification,squires2020active,choo2023subset}.
For each of the graph classes and parameters, we generate 100 DAGs and plot the average with an error bar.

\begin{enumerate}
    \item Erd\H{o}s-R\'{e}nyi styled graphs (used by \cite{squires2020active,choo2022verification})\\
    These graphs are parameterized by 2 parameters: number of nodes $n$ and density $\rho$.
    Generate a random ordering $\sigma$ over $n$ vertices.
    Then, set the in-degree of the $n^{th}$ vertex (i.e.\ last vertex in the ordering) in the order to be $X_n = \max\{1, \texttt{Binomial}(n-1, \rho)\}$, and sample $X_n$ parents uniformly form the nodes earlier in the ordering.
    Finally, chordalize the graph by running the elimination algorithm of \cite{koller2009probabilistic} with elimination ordering equal to the reverse of $\sigma$.\\
    \textbf{Parameters used:} $n = \{10, 15, 20, \ldots, 95, 100\}$ and $\rho = 0.1$.
    \item Tree-like graphs (used by \cite{squires2020active,choo2022verification})\\
    These graphs are parameterized by 4 parameters: number of nodes $n$, degree $d$, $e_{\min}$, and $e_{\max}$.
    First, generate a complete directed $d$-ary tree on $n$ nodes.
    Then, add $\texttt{Uniform}(e_{\min}, e_{\max})$ edges to the tree.
    Finally, compute a topological order of the graph by DFS and triangulate the graph using that order.
    As the original definition of this graph class by \cite{squires2020active} becomes very sparse as $n$ grows, we tweaked the other parameters to scale accordingly by defining new parameters $d_{prop}, e_{\min, prop}, e_{\max, prop} \in [0,1]$ as follows: $d = n \cdot d_{prop}$, $e_{\min} = n \cdot e_{\min, prop}$, and $e_{\max} = n \cdot e_{\max, prop}$.\\
    \textbf{Parameters used:} $n = \{100, 150, 200, \ldots, 450, 500\}$, $d_{prop} = 0.4$, $e_{\min, prop} = 0.2$, $e_{\max, prop} = 0.5$.
    \item $G(n,p)$-union-tree (used by \cite{choo2023subset})\\
    These graphs are parameterized by 2 parameters: number of nodes $n$ and edge probability $p$.
    An Erd\H{o}s-R\'{e}nyi $G(n,p)$ and a random tree $T$ on $n$ vertices are generated.
    Take the union of their edge sets, orient the edges in an acyclic fashion, then add arcs to remove v-structures.\\
    \textbf{Parameters used:} $n = \{10, 15, 20, \ldots, 95, 100\}$ and $p=0.03$.
\end{enumerate}

\subsection{Algorithms benchmarked}

While both the algorithm of \cite{choo2022verification} and Algorithm 2 have been implemented to take in a parameter $k$ for bounded-size interventions, our experiments focused on the case of atomic interventions, i.e.\ $k = 1$.

\texttt{separator}:\quad
Algorithm of \cite{choo2022verification}. With checks, it allows for full adaptivity.

\texttt{separator\_no\_check}:\quad
\texttt{separator} but we remove checks that avoid redundant interventions, i.e.\ $\cO(\log n)$ rounds of adaptivity.

\texttt{adaptive\_r1}:\quad
Algorithm 2 with $r = 1$, i.e.\ non-adaptive

\texttt{adaptive\_r2}:\quad
Algorithm 2 with $r = 2$

\texttt{adaptive\_r3}:\quad
Algorithm 2 with $r = 3$

\texttt{adaptive\_rlogn}:\quad
Algorithm 2 with $r = \log_2 n$

\texttt{adaptive\_r2logn}:\quad
Algorithm 2 with $r = 2 \log_2 n$. Can perform checks that avoid redundant interventions.

\texttt{adaptive\_r3logn}:\quad
Algorithm 2 with $r = 3 \log_2 n$. Can perform checks that avoid redundant interventions.

\texttt{adaptive\_rn}:\quad
Algorithm 2 with $r = n$, i.e.\ full adaptivity allowed

\subsection{Experimental results}

As expected, we observe that higher rounds of adaptivity leads to lower number of interventions required.
When $r \in \cO(\log n)$, Algorithm 2 can match \cite{choo2022verification} with checks disabled.
When $r=n$, Algorithm 2 can match \cite{choo2022verification} with its full adaptivity.

\begin{figure}[htb]
\centering
\begin{subfigure}[t]{\linewidth}
    \centering
    \includegraphics[width=\linewidth]{exp1_interventioncount.png}
    \caption{Number of interventions}
\end{subfigure}
\\
\begin{subfigure}[t]{\linewidth}
    \centering
    \includegraphics[width=\linewidth]{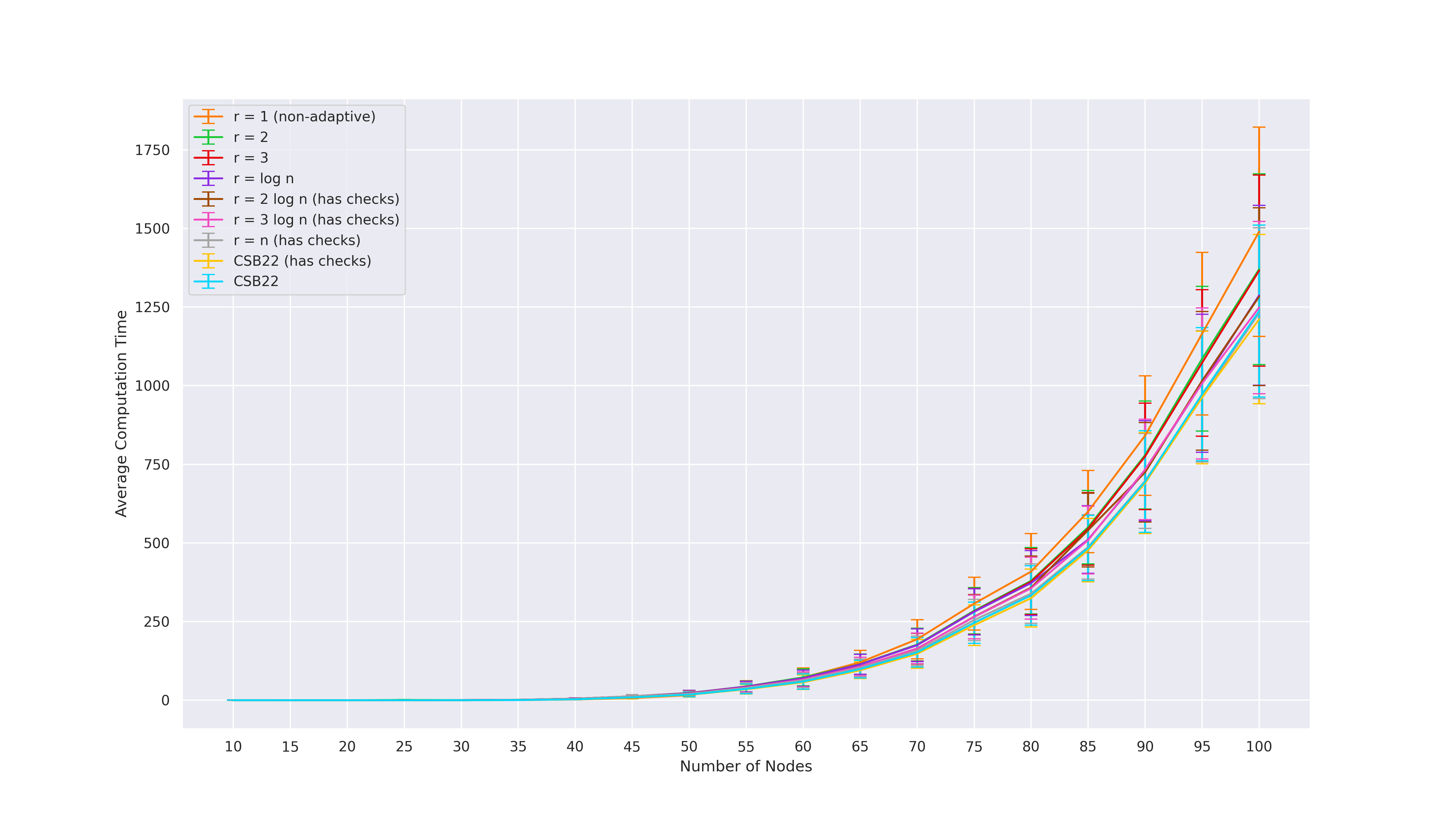}
    \caption{Time}
\end{subfigure}
\caption{Experiment 1}
\label{fig:exp1}
\end{figure}

\begin{figure}[htb]
\centering
\begin{subfigure}[t]{\linewidth}
    \centering
    \includegraphics[width=\linewidth]{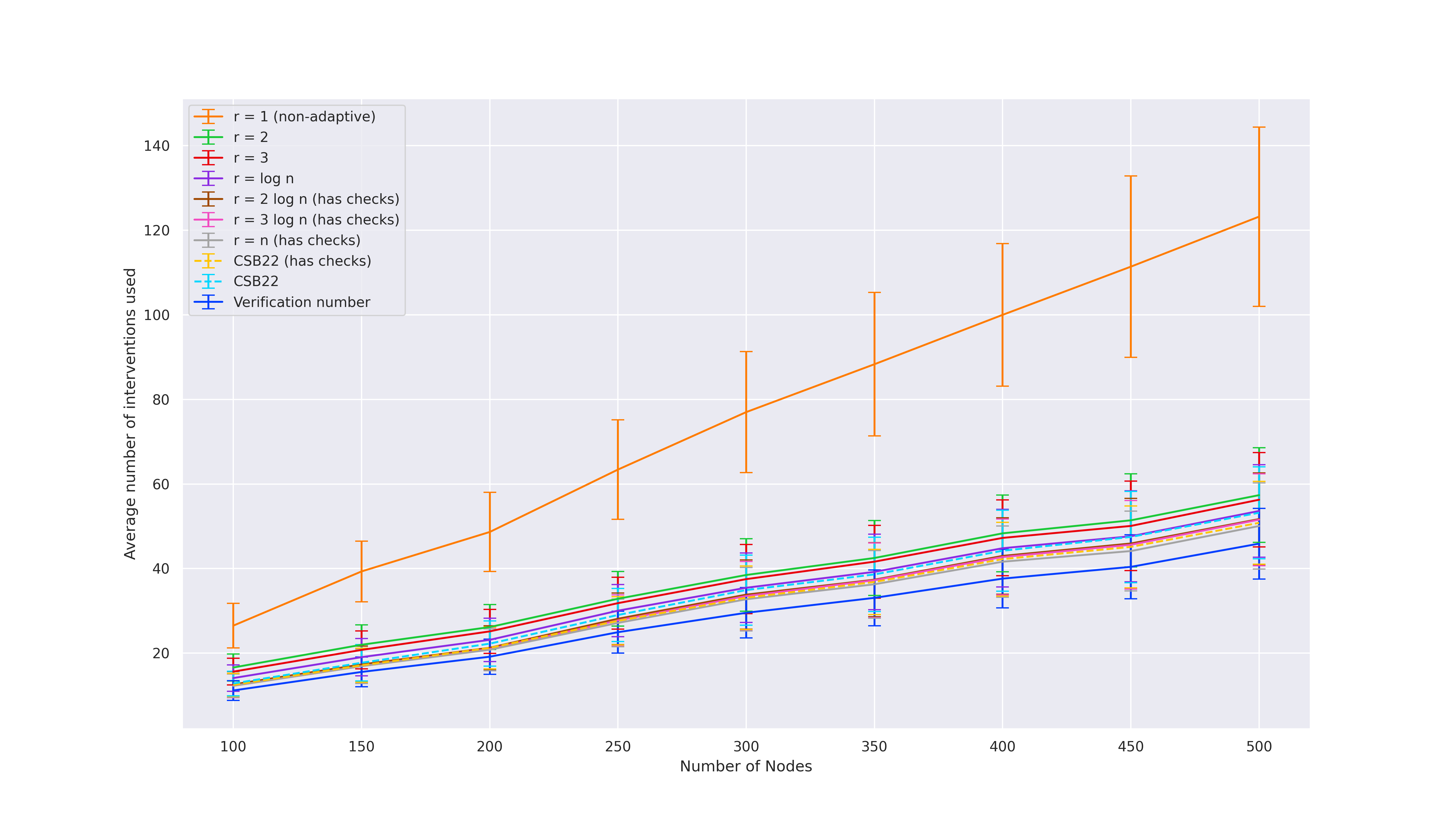}
    \caption{Number of interventions}
\end{subfigure}
\\
\begin{subfigure}[t]{\linewidth}
    \centering
    \includegraphics[width=\linewidth]{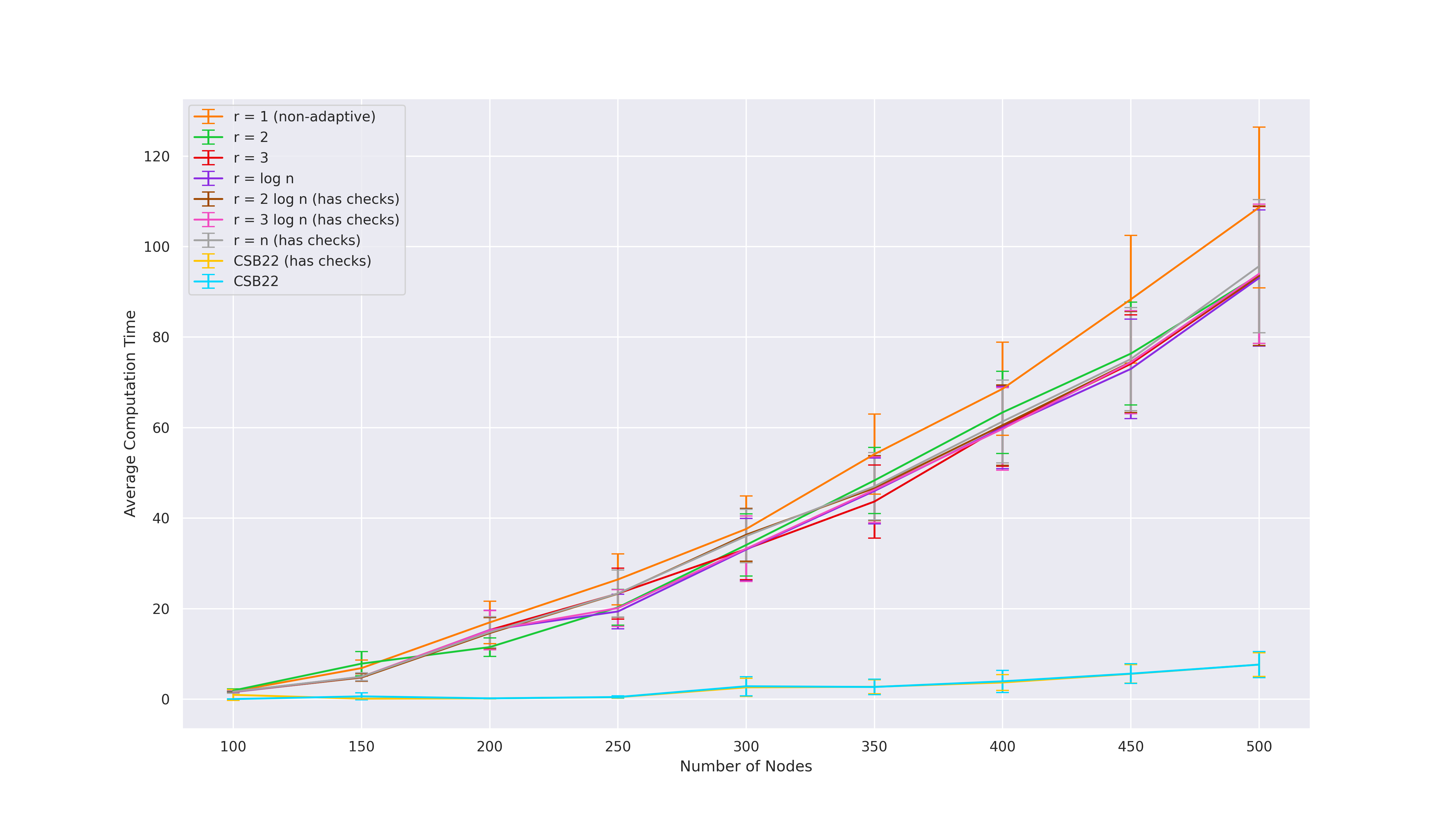}
    \caption{Time}
\end{subfigure}
\caption{Experiment 2}
\label{fig:exp2}
\end{figure}

\begin{figure}[htb]
\centering
\begin{subfigure}[t]{\linewidth}
    \centering
    \includegraphics[width=\linewidth]{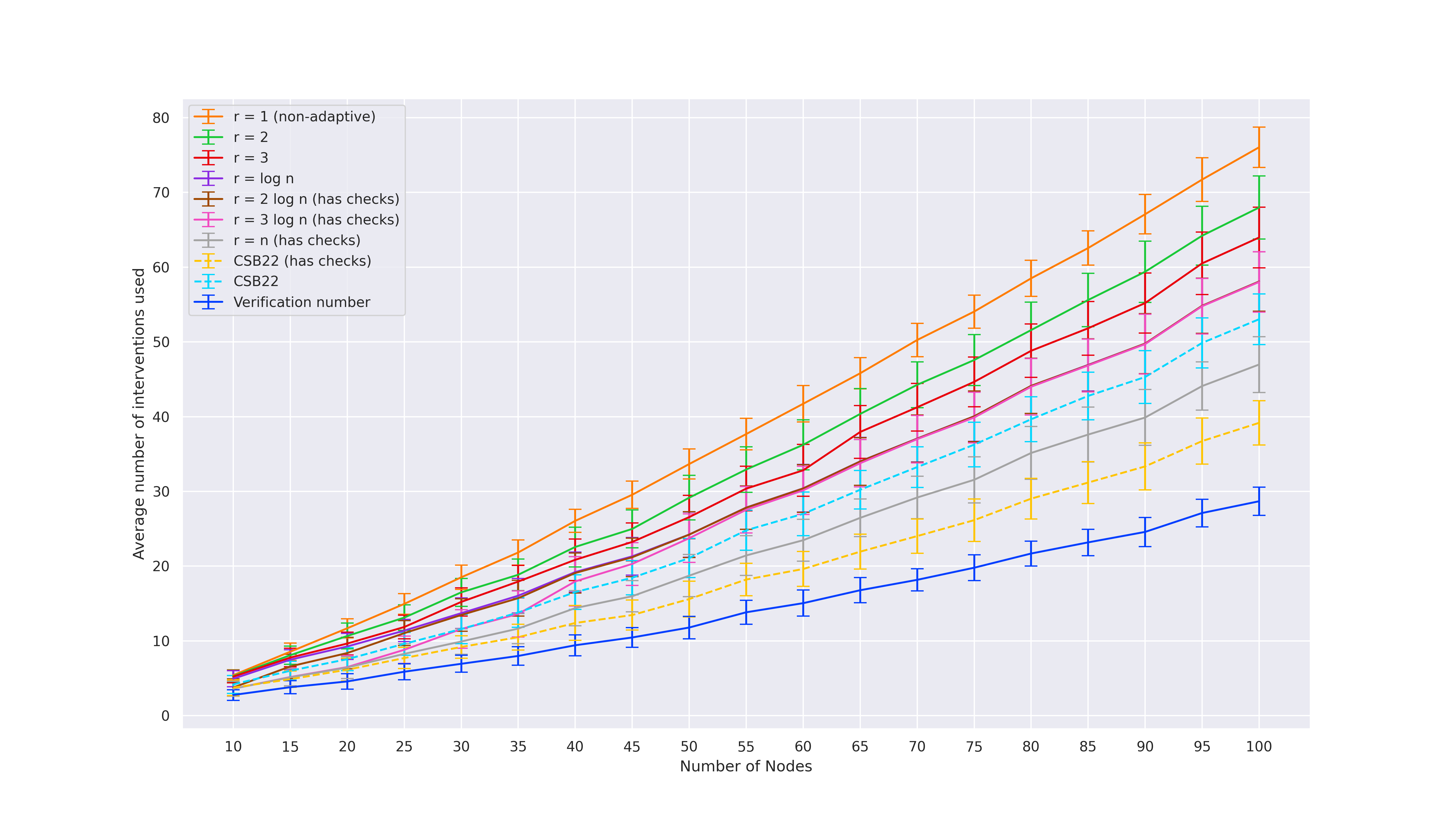}
    \caption{Number of interventions}
\end{subfigure}
\\
\begin{subfigure}[t]{\linewidth}
    \centering
    \includegraphics[width=\linewidth]{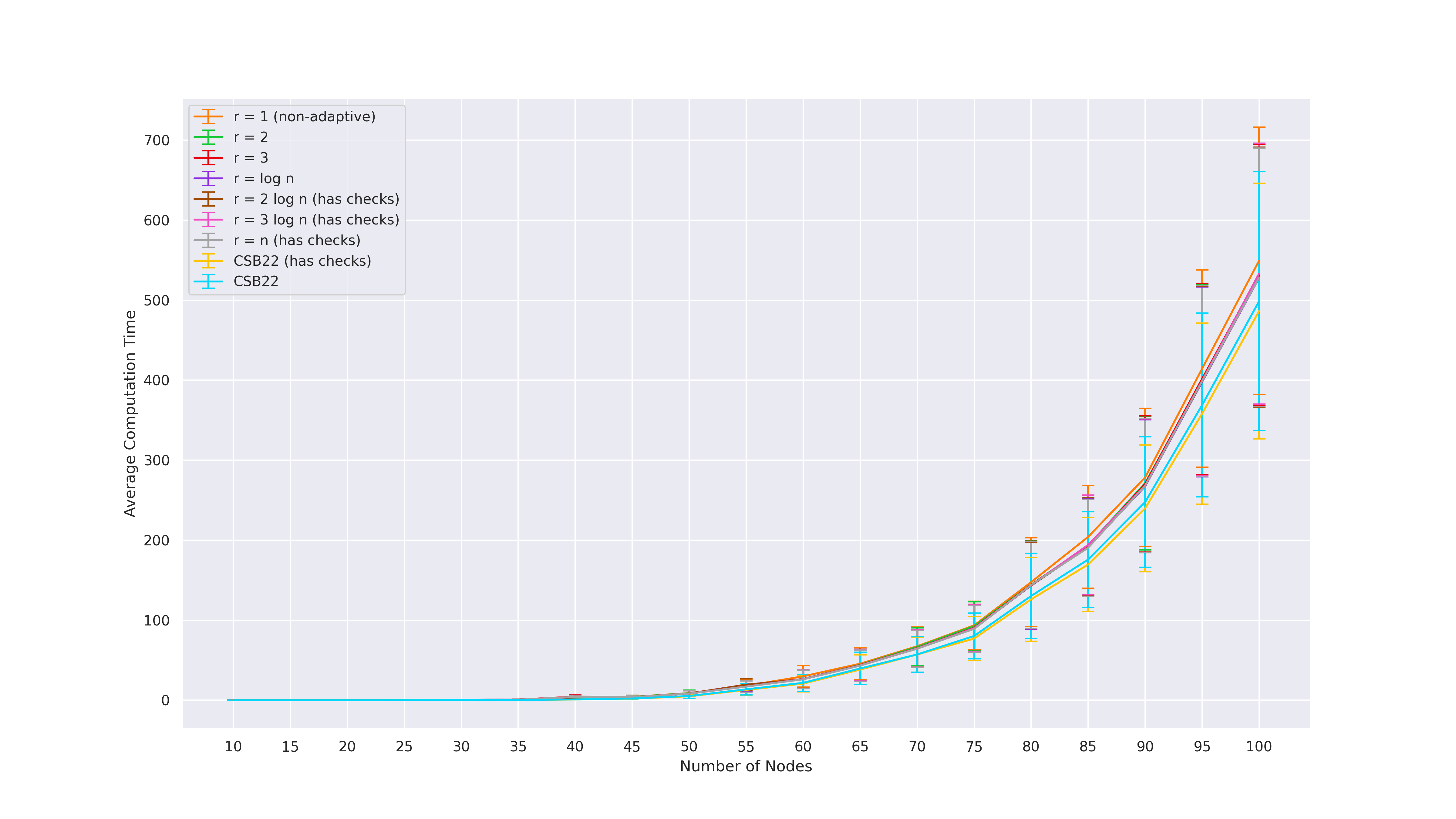}
    \caption{Time}
\end{subfigure}
\caption{Experiment 3}
\label{fig:exp3}
\end{figure}

\end{document}